\documentclass[11pt]{article}
\usepackage[preprint]{acl}
\usepackage{booktabs}
\usepackage{multirow}
\usepackage{tabularx}
\usepackage{times}
\usepackage{latexsym}
\usepackage[T1]{fontenc}
\usepackage[utf8]{inputenc}
\usepackage{microtype}
\usepackage{inconsolata}
\usepackage{subcaption}
\usepackage{graphicx}
\usepackage{booktabs}
\usepackage{amsmath}
\usepackage{tcolorbox}
\usepackage{hyperref}

\title{Rational Synthesizers or Heuristic Followers? Analyzing LLMs in RAG-based Question-Answering}

\author{Atharv Naphade \\ Carnegie Mellon University \\
  \texttt{anaphade@andrew.cmu.edu}}

\begin{document}
\maketitle

\begin{abstract}
Retrieval-Augmented Generation (RAG) is the prevailing paradigm for grounding Large Language Models (LLMs), yet the mechanisms governing \textit{how} models integrate groups of conflicting retrieved evidence remain opaque. Does an LLM answer a certain way because the evidence is factually strong, because of a prior belief, or merely because it is repeated frequently? To answer this, we introduce \textbf{GroupQA}, a curated dataset of 1,635 controversial questions paired with 15,058 diversely-sourced evidence documents, annotated for stance and qualitative strength. Through controlled experiments, we characterize group-level evidence aggregation dynamics: Paraphrasing an argument can be more persuasive than providing distinct independent support; Models favor evidence presented first rather than last, and Larger models are increasingly resistant to adapt to presented evidence. Additionally, we find that LLM explanations to group-based answers are unfaithful.  Together, we show that LLMs behave consistently as vulnerable heuristic followers, with direct implications for improving RAG system design.
\end{abstract}

\section{Introduction}

The integration of external knowledge via Retrieval-Augmented Generation (RAG) has become the standard solution for mitigating hallucinations in Large Language Models (LLMs) \citep{lewis2020retrieval, guu2020retrieval}. By retrieving relevant documents and placing them into the context window, RAG systems aim to ground model outputs in verifiable facts. This paradigm relies on an implicit assumption: that LLMs act as rational aggregators capable of weighing conflicting evidence to synthesize a coherent truth \citep{zhang2023siren}.

Prior benchmarks, such as \textit{ConflictingQA} \citep{wan-etal-2024-evidence}, have studied how models handle pairwise conflicts (one positive vs. one negative document). However, real-world retrieval is rarely a clean one-to-one comparison. A search for a contested topic typically returns a noisy top-$k$ list containing clusters of evidence: perhaps five documents supporting one claim, three supporting another, and various redundancies spread throughout. This leaves a critical gap in our understanding: how do LLMs behave when faced with \textit{groups} of potentially conflicting evidence?

We introduce \textbf{GroupQA}, a dataset explicitly designed to mimic the dynamics of real retrieval scenarios. Unlike previous benchmarks, GroupQA targets controversial binary questions paired with dense clusters of documents (avg. 9.2 documents per question), allowing for the manipulation of groups of documents rather than just individual documents. This allows us to isolate and study vital features of documental influence —such as the quantity of documents, their ordering, and their semantic diversity—to determine what actually drives an LLM's final belief.

Using GroupQA, we conduct a controlled analysis of state-of-the-art open and closed models. First, we show that larger models tend to incorporate documental evidence lesser. We then show that group structure fundamentally alters model behavior. We observe a potent \textit{Illusory Truth Effect}, where models decisions are changed more on average by repeating a single paraphrased document, rather than providing distinct evidence. Furthermore, we identify a persistent \textit{Primacy Effect}, where the first few documents in a group exert disproportionate influence on the final output. Finally, we benchmark the detection  These effects do not emerge from pairwise analysis and cannot be predicted by document-level comparisons alone. 

Our contributions are as follows:
\begin{enumerate}
    \item \textbf{The GroupQA Dataset:} We introduce GroupQA, a benchmark of \textbf{1635} controversial binary questions paired with \textbf{15,058} clustered retrieved documents that vary in stance,  enabling systematic evaluation of how LLMs respond to groups of contradictory information.  
    
\item \textbf{Persuasion Evaluation Methods}
We propose and evaluate models on simple, intervention-based metrics—including answer flip thresholds, belief plasticity, and leave-one-out document influence—to characterize how evidence quantity, diversity, ordering, and conflict affect model decisions across scales.

\item \textbf{Group Dynamics}
We show that group structure has significant influence on model answers: repetition of a single evidence increases its persuasiveness and can outweigh diverse evidence sets, model scale trades off belief flexibility and stability, explicitly conflicting evidence stabilizes decisions, and the order of evidence presented impacts evidence favor-ability. 
\end{enumerate}

\section{Related Work and Motivation}

\paragraph{Conflicting Information in RAG.}
Handling contradictory retrieval has become critical as RAG systems deploy to open-domain settings. \citet{longpre2021entity} exposed entity-based knowledge conflicts, finding models rely on parametric memory when passages contradict. \citet{chen2022rich} showed retrieval-based LLMs depend on non-parametric evidence when recall is high, but confidence scores fail to reflect inconsistencies. Recent work developed conflict taxonomies \citep{wang2025accommodate} and conflict-aware fine-tuning methods \citep{gao2024probing}. \citet{wan-etal-2024-evidence} analyzed pairwise document synthesis in \textit{ConflictingQA}. GroupQA provides document sets to study group-level dynamics like accumulation and majority voting.

\paragraph{Sycophancy and Contextual Persuasion.}
LLMs trained with RLHF exhibit susceptibility to user influence. \citet{sharma2023understanding} showed models conform to user-stated views even when incorrect and second-guess correct answers when users express doubt. \citet{wei2023simple} extended this to persona-based prompting. Our work investigates contextual persuasion—where influence comes from retrieved document composition rather than explicit user statements.

\paragraph{The Illusory Truth Effect in LLMs.}
In psychology, repeated statements are rated as more truthful \citep{hasher1977frequency}. \citet{min2022rethinking} found evidence of similar behavior in models, suggesting in-context learning is driven by label distribution over task understanding. We quantify this effect in RAG, measuring how paraphrased repetitions override parametric priors and showing redundancy can outweigh informational diversity.

\paragraph{Attention and Ordering Biases.}
\citet{liu2023lost} identified "Lost in the Middle," where performance degrades for information in long context centers, exhibiting a U-shaped curve. \citet{peysakhovich2023attention} and \citet{xiao2023efficient} attributed this to attention sinks weighting initial tokens. Our primacy bias findings extend these to behavioral evidence aggregation, showing early evidence establishes anchors that subsequent reasoning fails to override.

\paragraph{Multi-Document Reasoning.}
Recent work examined information aggregation across documents. \citet{zhang2023siren} surveyed hallucination mitigation including conflicting source synthesis, \citet{yoran2024making} developed methods for retrieval robustness, and \citet{xu2024knowledge} studied parametric-nonparametric knowledge balancing under conflict. Our work systematically characterizes the heuristics models use, providing a foundation for understanding what must be overcome.

\section{The GroupQA Dataset}
\label{sec:dataset}

We now describe the construction of \textbf{GroupQA}, our dataset designed to evaluate what types of evidence sets influence LLM decisions. We designed GroupQA to emulate the common setup for deploying retrieval-augmented LLMs: we retrieve the most relevant documents for a particular user query and place them in the LLM’s context window. To build our dataset, we tackle three challenges: collecting contentious binary questions, identifying relevant and diverse evidence paragraphs from the open web, and filtering for contentious sets of documents for each question. All prompts used in this section are specified in Appendix \ref{app:dataset_details}

\paragraph{Collecting contentious questions.}
We first create a series of realistic open-ended questions for which there exists conflicting evidence online. Critically, unlike past work on ambiguity in QA, we aim to collect unambiguous questions that still have answer conflicts due to societal debate or common misconceptions. We design the questions to elicit binary responses of \textit{Yes} or \textit{No} to simplify evaluation.
\begin{table}[t!]
    \centering
    \small
    
    \resizebox{\columnwidth}{!}{%
    \begin{tabular}{lr}
        \toprule
        \textbf{Metric} & \textbf{Value} \\
        \midrule
        Avg. words per paragraph& 384.4 \\
        Avg. paragraphs per Q & 9.21 \\
        Avg. Yes / No paragraphs & 4.54 / 4.67 \\
        Questions with $\ge 1$ Strong Doc & 42.2\% \\
        Stance Skew (Avg Yes - Avg No) & -0.12 \\
        \bottomrule
    \end{tabular}%
    }
    \captionof*{table}{\textbf{(a) GroupQA Statistics}}

    \vspace{2ex} 

    \resizebox{\columnwidth}{!}{%
    \begin{tabular}{lrrrr}
        \toprule
        \multicolumn{5}{l}{\textbf{Collection \& Filtering}} \\
        \midrule
        \multicolumn{2}{l}{Questions Processed: 1,883} & \multicolumn{3}{r}{Questions Accepted: 1,635} \\
        \multicolumn{2}{l}{Total Docs Scraped: 22,264} & \multicolumn{3}{r}{Acceptance Rate: 86.83\%} \\
        \midrule
        \multicolumn{5}{l}{\textbf{Evidence Distribution}} \\
        \midrule
        \textbf{Stance} & \textbf{Strong} & \textbf{Medium} & \textbf{Weak} & \textbf{Total} \\
        \midrule
        Yes & 1,019 & 1,815 & 4,594 & 7,428 \\
        No  & 394   & 504   & 6,732 & 7,630 \\
        \midrule
        \textbf{Total} & \textbf{1,413} & \textbf{2,319} & \textbf{11,326} & \textbf{15,058} \\
        \bottomrule
    \end{tabular}%
    }
    \captionof*{table}{\textbf{(b) Data Collection \& Distribution}}
    
    \vspace{1ex}
    \caption{Comprehensive GroupQA Statistics. \textbf{(a)} The dataset is dense and balanced. \textbf{(b)} Details of the scraping pipeline and evidence distribution.}
    \label{tab:combined_stats}
\end{table}

We create questions using GPT-4o. To ensure that the model generates a diverse set of questions, we take inspiration from previous work in synthetic dataset generation and stratify the generations by topic: we first generate 95 distinct categories (e.g., \textit{Bioengineering, Zoology, Historical Revisionism}), then generate sets of questions conditioned on each category. We qualitatively confirm that the questions are diverse and challenging; specific examples include \textit{``Does caffeine improve long-term memory?''} and \textit{``Is nuclear power considered renewable?''}. In addition, we manually remove duplicate questions using cosine similarity filtering (threshold $> 0.92$).

This process yielded an initial pool of \textbf{1,948 candidate questions}.

\paragraph{Collecting evidence paragraphs.}
Given these questions, we want to find evidence paragraphs that support both the answers of \textit{Yes} and \textit{No}. To handle this, we emulate running a real-world retrieval-augmented LLM system that uses the Google Search API as its retrieval engine.

We first turn each question into affirmative and negative assertions using a deterministic template. For example, the question \textit{``Do vaccines cause autism?''} is converted to \textit{``Vaccines cause autism''} ($A_{pos}$) and \textit{``Vaccines do not cause autism''} ($A_{neg}$). For both positive and negative statements, we search the queries using the Google Search API and retrieve the top documents $k$ (where $k=10$). 

As is common in many retrieval-augmented models, we do not consider visual features. Instead, we extracted the raw text from each document using the \texttt{Trafilatura} library, which we found superior to standard HTML parsers for boilerplate removal. Additionally, we do not explicitly include metadata like source URL or publication date to force the model to rely solely on textual content.

\paragraph{Filtering and Stance Labeling.}
When searching queries such as \textit{``vaccines cause autism''}, we inevitably retrieve documents that argue the opposite or are irrelevant. To label the actual position of the document, we use GPT-4o-mini and Gemini-2.5-Flash as a judge (Both must agree or we scrap the question). We prompt the model to classify each document’s stance (Affirmative, Negative, or Neutral) and assign a qualitative strength score (Strong, Medium, or Weak) based on the presence of citations or expert testimony. The strength scores are not used in our experiments due to their subjectivity. To ensure accuracy of stance labeling, we sampled 100 random documents and manually verified 99\% of ratings.

From the initial 1,948 questions, we filter out any question where we could not retrieve at least one valid supporting document for \textit{both} sides of the argument. This resulted in the removal of 313 questions (16.1\% attrition). The final \textbf{GroupQA} dataset consists of \textbf{1,635 questions} paired with \textbf{15,058 evidence documents}.

\paragraph{Creating conflicting examples.}
Finally, we want to isolate paragraphs from these larger documents to feed into the LLMs. To do this, we extract the most relevant paragraph from each document. We run the \texttt{all-MiniLM-L6-v2} model to embed both the question and all candidate paragraphs, computing the cosine similarity to select the highest-scoring window. Table~\ref{tab:combined_stats} presents the basic statistics for our final data. While ``Weak'' evidence is most common (reflecting the nature of the open web), 42.2\% of questions contain at least one ``Strong'' document

\begin{table*}[t]
\centering
\footnotesize
\caption{Dataset Sample:  Affirmative vs. Negative Evidence Snippets}
\label{tab:diamond-dataset-final}
\begin{tabularx}{\textwidth}{p{0.16\textwidth} X X}
\toprule
\textbf{Question} & \textbf{Affirmative Evidence (Yes)} & \textbf{Negative Evidence (No)} \\
\midrule
\multirow{22}{=}{\textbf{Does the diamond industry contribute positively to economic development in mining regions?}} 
& \textbf{[Strength: STRONG]} \newline The diamond mining industry is a major driver of economic growth in many countries, particularly in Africa, where some of the world's largest diamond reserves are located. The diamond industry contributes \$16 billion in total net economic benefits annually... ... 
& \textbf{[Strength: STRONG]} \newline In many conflict-prone regions, mining activities often contribute to both environmental degradation and the intensification of local conflicts. These issues are exacerbated by weak governance structures, poor enforcement of regulations, corruption and limited accountability... ... \\
\cmidrule(lr){2-3}
& \textbf{[Strength: WEAK]} \newline Economic Contributions Job Creation in Diamond Mining Regions The diamond mining industry is a significant source of employment, particularly in regions where economic opportunities may be limited. In countries like Botswana and South Africa, diamond mining has created thousands... ... 
& \textbf{[Strength: WEAK]} \newline across mineral-rich West African countries. Their involvement in legal and illegal large-scale corporate mining, as well as the small-scale artisanal mining intended for locals, demands policy responses that strengthen governance and enforcement across the region. ... \\
\cmidrule(lr){2-3}
& \textbf{[Strength: WEAK]} \newline This is particularly true as we continue to strengthen critical minerals supply and promote innovation and sustainable practices across critical minerals value chains. We are doing this in a way that supports regional economic growth; creates a more inclusive... ... 
& \textbf{[Strength: WEAK]} \newline Hilson, G. and S. Van Bockstael, 2012: Poverty and livelihood diversification in rural Liberia: exploring the linkages between artisanal diamond mining and smallholder rice production. Journal of Development Studies, 48 (3), 413–428, doi: https://doi.org/10.1080/00220388.2011.604414. ... \\
\cmidrule(lr){2-3}
& \textbf{[Strength: WEAK]} \newline When it comes to value-added activities, Botswana, Senegal and South Africa have established strong frameworks to encourage local mineral processing and beneficiation. Botswana has also made significant strides in value addition, particularly in the diamond sector... ... 

\end{tabularx}
\end{table*}

\section{Experimental Results}
\label{sec:results}

In this section, we use GroupQA to evaluate how models make conflicting decisions with documents. 

\subsection{Model Answers}
We define the model's answer space as a probability distribution over the binary options $\mathcal{Y} = \{\text{``Yes''}, \text{``No''}\}$. We may refer to this as its belief or decision. For a parameterized model $\theta$ and a question $q_i$, we define the \textit{prior} belief as the normalized probability assigned to the ``Yes'' token:
\begin{equation}
    P_{\text{prior}} = P_{\theta}(\text{Yes} \mid q_i)
\end{equation}
To ensure a valid probability distribution, we normalize the model's output probabilities for ``Yes'' and ``No'' such that they sum to 1 (exact prompts and normalization details are provided in the Appendix).

To measure the effect of evidence, we condition the model on the set of supporting documents $D_i = \{d_{i,1}, \dots, d_{i,m}\}$ associated with $q_i$. We refer to this as the \textit{posterior} decision:
\begin{equation}
    P_{\theta}(\text{Yes} \mid q_i, D_i)
\end{equation}
where the model context includes the concatenation of $q_i$ and the evidence set $D_i$.

Table~\ref{tab:aggregate-results} presents the aggregate results. Llama-3.1-70B's decisions show a positive correlation with the majority viewpoint; specifically, the model's output aligned with the majority perspective in $69\%$ of cases when conditioned on the full evidence set $D_i$. The posterior decision was only $0.074 \pm 0.084$ more than the prior with $95\%$ confidence.

\paragraph{Robustness of Model Answers.}
To ensure our findings are robust to distinct prompting methods,  we compute the mean probability shift between standard and CoT prompting: $\Delta_{\text{CoT}} =  P_{\text{CoT}}(\text{Yes}) - P_{\text{Standard}}(\text{Yes})$. 

\begin{figure}[t!]
    \centering
    \includegraphics[width=\linewidth]{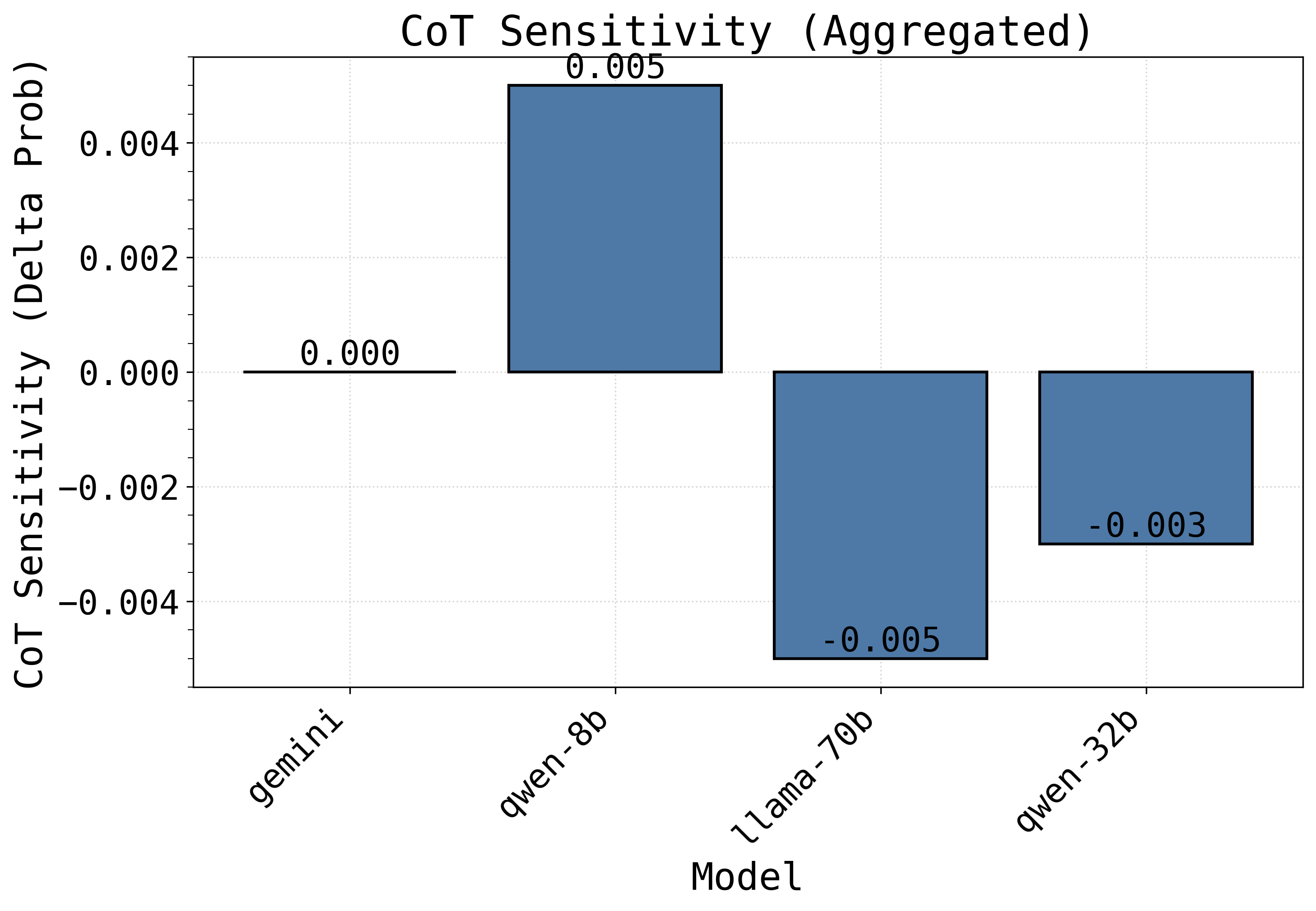}
    \caption{Aggregate shift in belief probability before and after CoT. The negligible delta suggests reasoning does not significantly alter the underlying belief distribution.}
    \label{fig:cot_impact}
\end{figure}

As shown in Figure~\ref{fig:cot_impact}, the impact of CoT is negligible across all model scales, resulting in an absolute probability mass shift of $< 0.5 \pm 0.56\%$ with 95\% confidence. This suggests that for this domain, reasoning traces function primarily as post-hoc rationalizations rather than corrective inference steps. Consequently, our findings on belief dynamics hold robustly across both standard and reasoning-augmented generation.

\subsection{Attribution Unfaithfulness}
\citet{wan-etal-2024-evidence} demonstrated that models often fail to verbally estimate the persuasive weight of evidence. We investigate this phenomenon in the context of multi-document reasoning, as the explainability of agentic decision-making is critical for safety. We compare the model's \textit{verbalized importance} against the true \textit{causal importance} of each document.

For a subset of 200 questions, we presented the model with the full set of permuted documents $D_i$ and prompted it to identify the index $r_i$ of the document that, if removed, would alter its decision the most. We contrast this with the ground-truth causal importance, determined via a Leave-One-Out perturbation analysis. We identify the document $d_{i,k}$ that, when removed, maximizes the divergence from the original belief state:

\begin{equation}
\begin{split}
    k_i = \operatorname*{argmax}_{j} \Big| & P_{\theta}(\text{Yes} \mid q_i, D_i \setminus \{d_{i,j}\}) \\
    & - P_{\theta}(\text{Yes} \mid q_i, D_i) \Big|
\end{split}
\end{equation}

We define the model as \textit{faithful} if the verbalized attribution matches the causal reality ($r_i = k_i$). Our results indicate that Llama-3.1-70B is faithful on only $26\%$ of questions. This confirms that we cannot rely on self-reported attribution to determine document utility. Consequently, the remainder of our experiments utilize causal methods to diagnose model decision-making.

\begin{table}[t]
\centering
\small
\resizebox{\columnwidth}{!}{
\begin{tabular}{lcccc}
\toprule
\textbf{Model} & \textbf{Ent.} & \textbf{Prior $P(Y)$} & \textbf{All $P(Y)$} & \textbf{Maj} \\
\midrule
DeepSeek-R1-8B & 0.852 & 0.551 & 0.629 & 0.73 \\
Gemini-2.5-FL  & 0.762 & 0.640 & 0.641 & 0.68 \\
Llama-3.1-70B  & 0.758 & 0.681 & 0.688 & 0.69 \\
Qwen3-32B      & 0.843 & 0.607 & 0.671 & 0.70 \\
\bottomrule
\end{tabular}
}
\caption{Aggregate Statistics for Model Priors and Stability. \textbf{Ent.}: Entropy of Prior. \textbf{Prior $P(Y)$}: Avg. probability of 'Yes' before evidence. \textbf{All $P(Y)$}: Avg. probability of 'Yes' with full evidence. \textbf{Maj}: Probability of agreement with majority.}
\label{tab:aggregate-results}
\end{table}

\subsection{Plasticity and Belief Stability}
We introduce the metric of \textit{Plasticity} ($PL_{\theta, D}$) to quantify the sensitivity of a model's prior beliefs to external evidence. It is defined as the mean absolute shift in probability mass assigned to the "Yes" token when conditioned on retrieved documents $D_i$ versus the parametric prior:

\begin{equation}
\begin{split}
    PL_{\theta, D} = \frac{1}{n}\sum_{i}^{n} \Big| & P_{\theta}(\text{Yes} \mid q_i, D_i) \\
    & - P_{\theta}(\text{Yes} \mid q_i) \Big|
\end{split}
\end{equation}

\begin{figure}[t!]
    \centering
    \includegraphics[width=\linewidth]{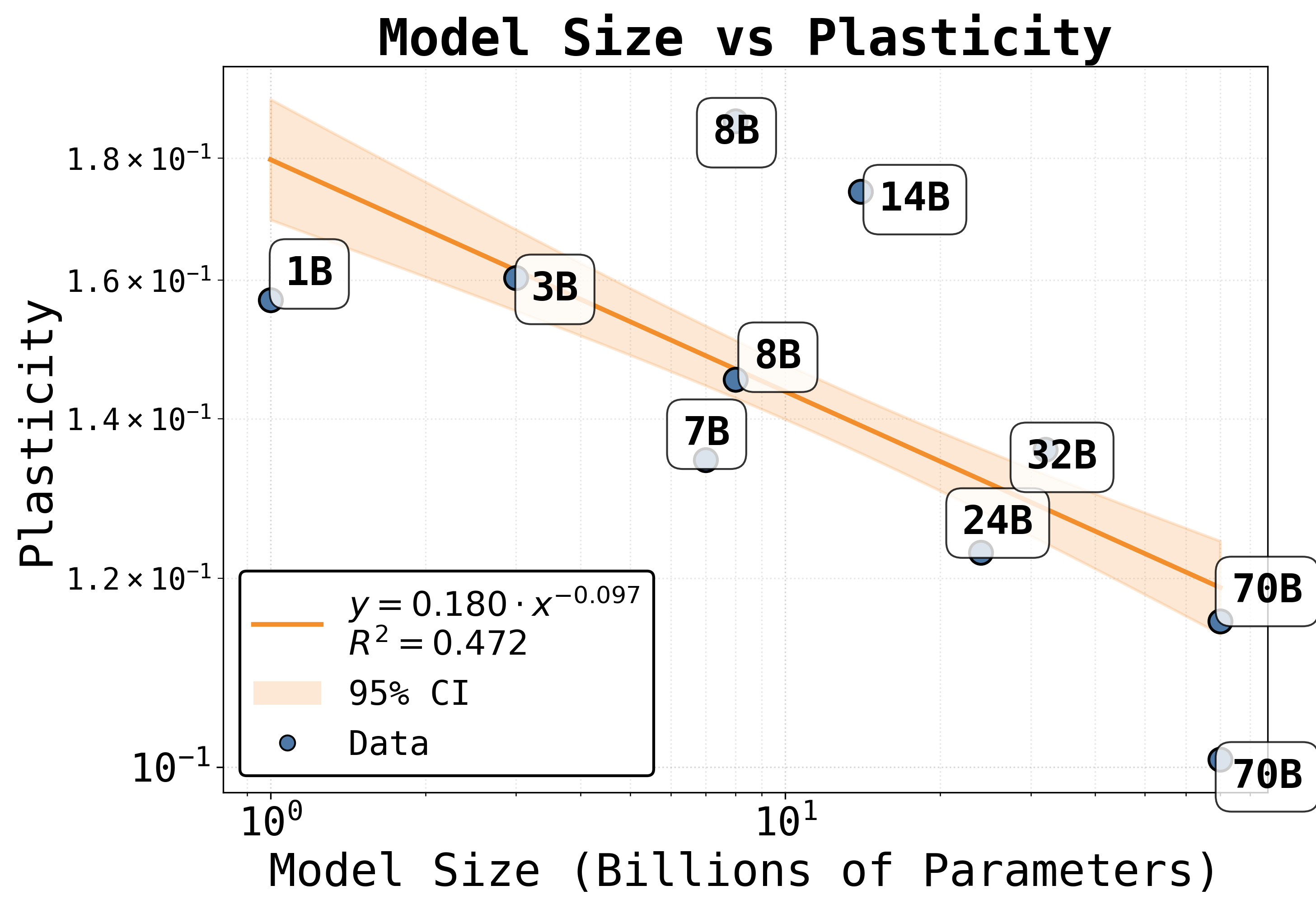}
    \caption{\textbf{Scaling Trend: Model Size vs. Plasticity.} Analysis of 10 open-weight models}
    \label{fig:scaling_law}
\end{figure}

\paragraph{Impact of Model Scale on Belief Plasticity.}
We observe a general inverse relationship between the model scale and the plasticity of belief. As illustrated in Figure~\ref{fig:scaling_law}, larger models tend to exhibit higher rigidity. For example, Llama-3.1-70B displays minimal plasticity ($PL = 0.0074$), whereas smaller models such as DeepSeek-R1-8B are approximately $10\times$ more volatile ($PL = 0.075$), shifting their probability distribution aggressively in the presence of a new context.

To formalize this observation, we analyzed 10 distinct checkpoints from the Llama and Qwen families (1B to 70B parameters) on a random subset of 100 tasks. Despite differences between families contributing to variance ($R^2 = 0.472$), we find that plasticity consistently decays as parameter count increases, approximated by the power-law fit:
\begin{equation}
    y = 0.180 \cdot x^{-0.097}
\end{equation}
where $x$ represents parameters in billions. This trend indicates that as models scale, their priors become increasingly persistent, requiring exponentially stronger evidence to displace. While specific architectural choices (e.g., Qwen vs. Llama) influence the baseline plasticity, the downward trajectory remains robust across model families.

\setlength{\tabcolsep}{3pt} 
\renewcommand{\arraystretch}{1.1} 

\begin{table*}[t]
\centering
\small
\begin{tabular*}{\textwidth}{@{\extracolsep{\fill}} l cc c cc cc @{}}
\toprule
& \multicolumn{2}{c}{\textbf{General Dynamics ($N \approx 1630$)}} & & \multicolumn{2}{c}{\textbf{Distinct (Opposing)}} & \multicolumn{2}{c}{\textbf{Paraphrased (Opposing)}} \\
\cmidrule{2-3} \cmidrule{5-6} \cmidrule{7-8}
\textbf{Model} & \textbf{Total Flips} & \textbf{Rate} & & \textbf{Rate} & \textbf{$X_{\min}$} & \textbf{Rate} & \textbf{$X_{\min}$} \\
\midrule
DeepSeek-R1-8B & 279 & 17.1\% & & 67.6\% & 1.52 & 76.5\% & 2.01 \\
Gemini-2.5-FL  & 351 & 21.5\% & & 63.7\% & 1.44 & 75.6\% & 2.24 \\
Llama-3.1-70B-Instruct  & 197 & 12.1\% & & 62.9\% & 1.27 & 69.8\% & 1.67 \\
Qwen3-32B      & 170 & 10.4\% & & 67.3\% & 1.34 & 73.7\% & 1.95 \\
\bottomrule
\end{tabular*}
\caption{Comprehensive Analysis of Answer Flipping Dynamics. \textbf{General Dynamics} shows overall stability ($N=1630$). \textbf{Condition Breakdown} details flip rates and thresholds ($X_{\min}$) under specific evidence types: Distinct vs. Paraphrased.}
\label{tab:comprehensive-flip-dynamics}
\end{table*}

\subsection{Quantity Dynamics}
We investigate how the quantity and nature of evidence modulate model answers. We do this by providing evidence sets that all oppose model priors. Specifically, we measure the marginal impact of adding supporting documents on the binary answer state. 
\paragraph{Answer Flipping.}
To establish a causal link between retrieval documents and model answers, we evaluate answer sensitivity across two distinct output modalities: (1) the continuous shift in probability mass assigned to the ``Yes'' token, and (2) the discrete inversion of the generated answer (binary label flipping) under greedy decoding.

While discrepancies between generative outputs and probabilistic scores are expected due to calibration errors \citep{liu2024fullece}, these metrics offer complementary views on answering under uncertainty. We posit that if a specific set of documents $D_i$ is sufficient to invert the model's discrete label (flip the answer), those documents are causally instrumental to the answering process, overriding the model's priors.

Additionally, we define the \textit{Flip Threshold}, $X_{\min}$, as the mean minimum number of documents required to invert a model's decision from its prior state. This is computed only on instances where the model undergoes a decision shift within the context window (up to 10 documents) given an opposing prior. To control for document-specific variance, metrics are averaged over a random subset of 200 questions. We contrast two experimental conditions:
\begin{itemize}
    \item \textbf{Informational Diversity (Distinct):} Accumulation of unique, distinct supporting documents.
    \item \textbf{Redundancy (Paraphrased):} Accumulation of rephrased variations of a single supporting document. We generate rephrases with GPT-4o shown in Appendix ~\ref{app:dataset_details}
\end{itemize}

\paragraph{Aggregate Flipping Dynamics.}
Table~\ref{tab:comprehensive-flip-dynamics} summarizes the propensity of each model to revise its answer. In the general setting, we observe significant variance; Qwen3-32B exhibits the lowest flip rate (only 10.4\%), while Gemini-2.5-FL has the highest flip rate (21.5\%).

However, the most critical insight emerges when comparing evidence types. Contrary to the intuition that diverse evidence provides a stronger signal, we find that \textbf{redundancy drives higher belief revision rates than informational diversity}. For example, DeepSeek-R1-8B flips in 67.6\% of cases when provided with distinct evidence, but this rate jumps to 76.5\% when exposed to paraphrased variations of a single document. This trend holds across all models, with Gemini-2.5-FL showing a massive increase from 63.7\% (Distinct) to 75.6\% (rephrased).

\paragraph{Decisiveness.}
While flip rates indicate \textit{how often} a model flips its answer,  the \textit{Flip Threshold} $X_{\min}$, indicates how rapidly a model flips, correlated with the human trait of decisiveness.  Llama-3.1-70B is then the most decisive; although it flips less frequently overall (12.1\%), when it does yield to distinct counter-evidence, it reaches a tipping point rapidly ($X_{\min}^{\text{dist}} = 1.27$). This contrasts with smaller models like DeepSeek-R1-8B ($X_{\min}^{\text{dist}} = 1.52$), possibly explained by the fact that it is slower at in-context learning 

\paragraph{The Illusory Truth Effect} 
Figure~\ref{fig:illusory_truth_llama} details the flip dynamics for Llama-3.1-70B (see Appendix~\ref{app:illusory_truth} for all models). We observe a distinct crossover in evidence efficacy. In the low-evidence regime ($1$--$2$ documents), distinct evidence (solid line) is strictly required to initiate belief revision. However, as the context grows, redundant evidence (dashed line) scales more aggressively. 

This confirms the statistical dominance of rephrased evidence observed in Table~\ref{tab:comprehensive-flip-dynamics}. The behavior mirrors the \textit{Illusory Truth Effect} \citep{hasher1977frequency}, suggesting that in long-context windows, LLMs conflate repetition with consensus. Rather than aggregating more distinct viewpoints to form a robust conclusion, the models appear susceptible to a frequency-based bias, where the sheer volume of repetition outweighs the quantity of distinct information.

\begin{figure}[t!]
    \centering
    \includegraphics[width=\linewidth]{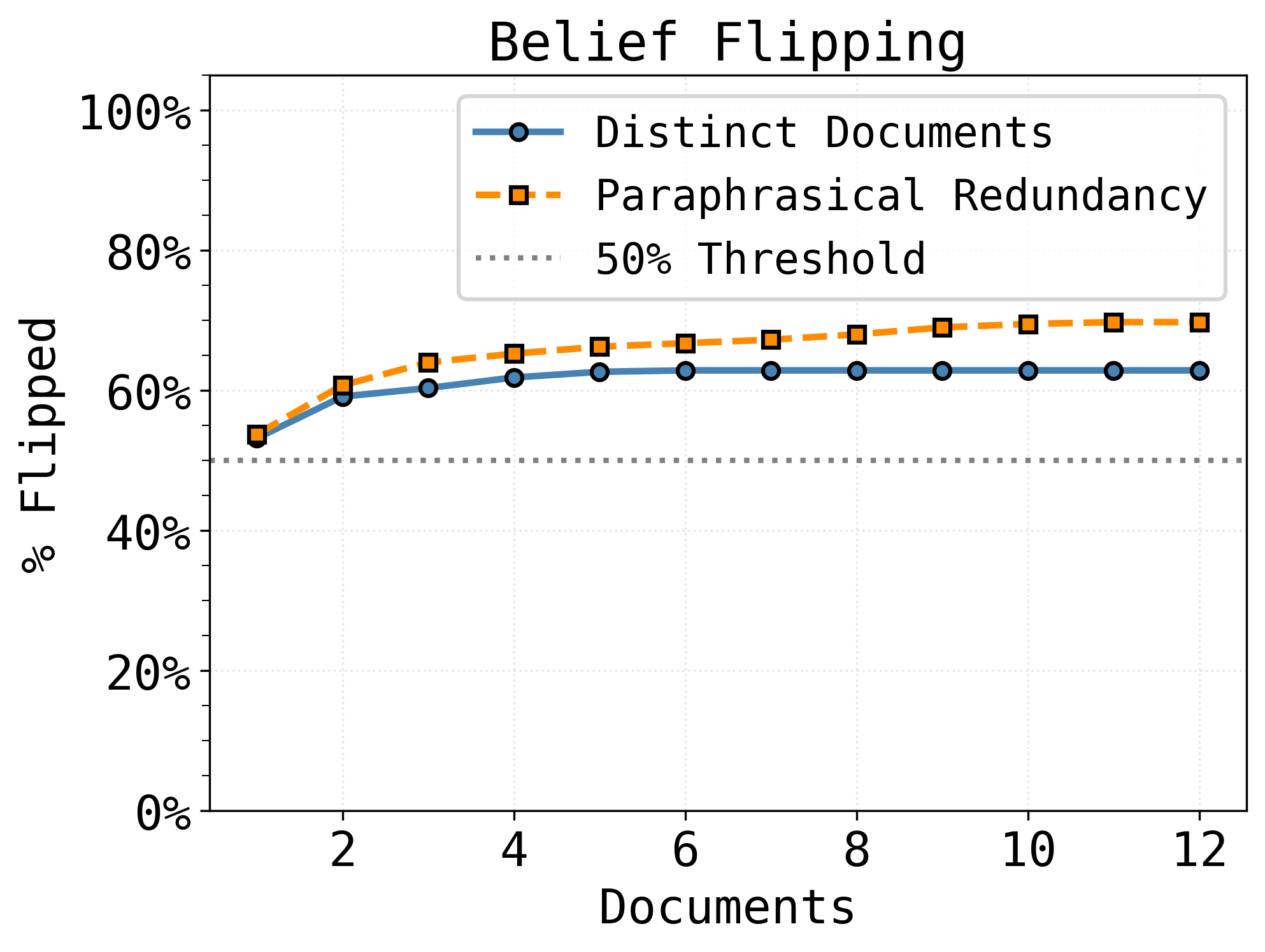}
    \caption{(Llama-3.1-70B-Instruct). Paraphrased vs distinct evidence flipping rate with all documents opposed to model prior.}
    \label{fig:illusory_truth_llama}
\end{figure}

\subsection{Dynamics under Conflicting Evidence}
\label{sec:conflict_dynamics}

Next, we examine belief updating when retrieved contexts contain explicitly conflicting information. Unlike prior experiments, where evidence challenged a static parametric prior, here we initialize the model with a \textit{balanced context} consisting of both supporting and opposing documents ($D_{\text{init}} = \{d_{\text{pos}}, d_{\text{neg}}\}$). This setting introduces active epistemic conflict before additional evidence is supplied.

\paragraph{Conflict Detection.}
To verify that models recognize and track this conflict, we measure both explicit conflict detection and document-level stance attribution accuracy. Across the models we examined, they detected knowledge conflicts over $89.8\%$ of the time, and correctly attributed each document to its stance over $76.2\%$ of the time. Empirically, this accuracy is especially low in highly one-sided document sets.

\begin{figure}[t!]
    \centering
    \includegraphics[width=\linewidth]{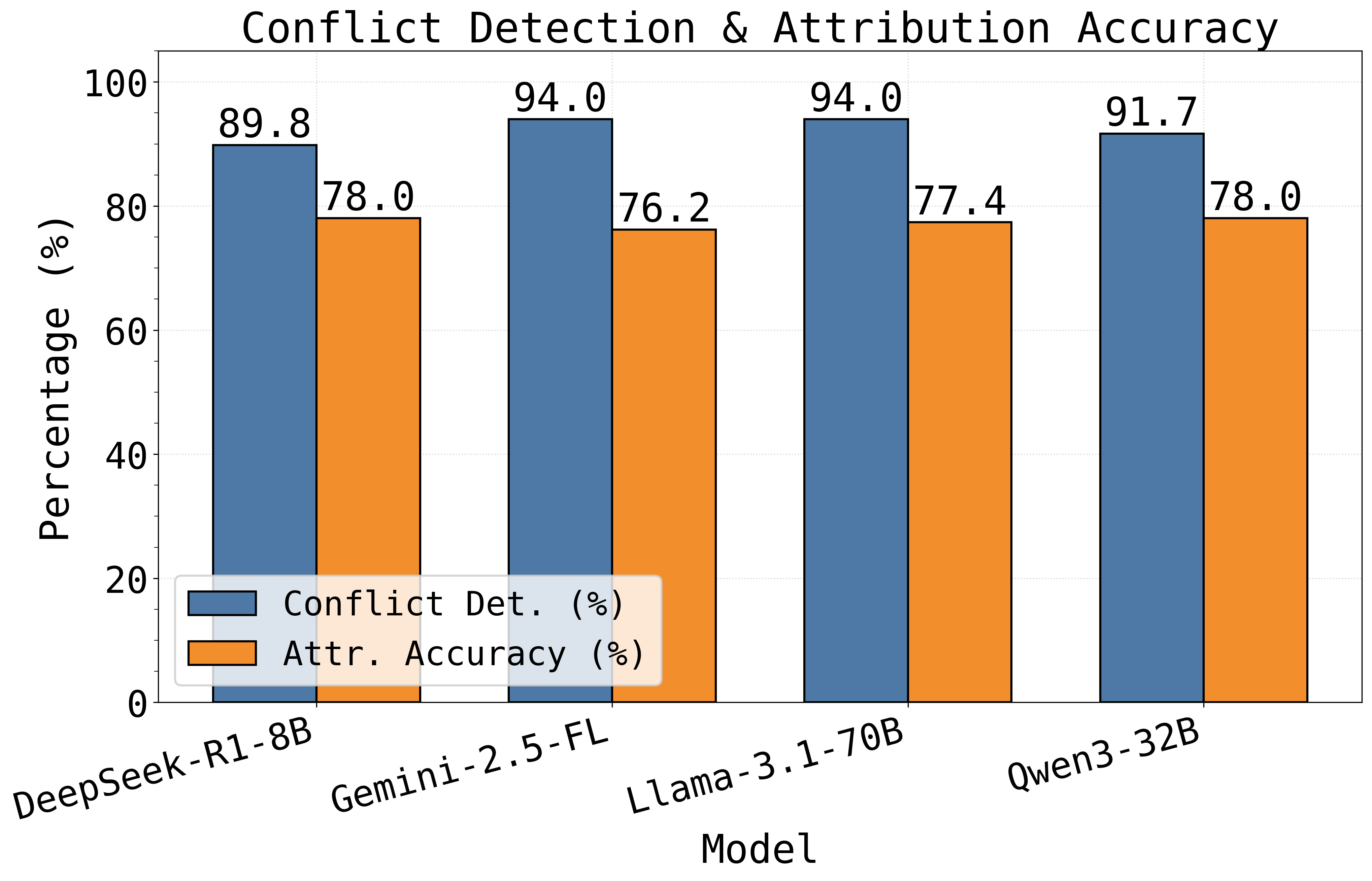}
    \caption{Conflict awareness metrics. Conflict Detection. measures the proportion of instances where the model explicitly flags a contradiction. Attribution Accuracy measures the precision of assigning the correct stance (Yes/No) to each retrieved document.}
    \label{fig:conflict_detection}
\end{figure}

\begin{figure}[t!]
    \centering
    \includegraphics[width=\linewidth]{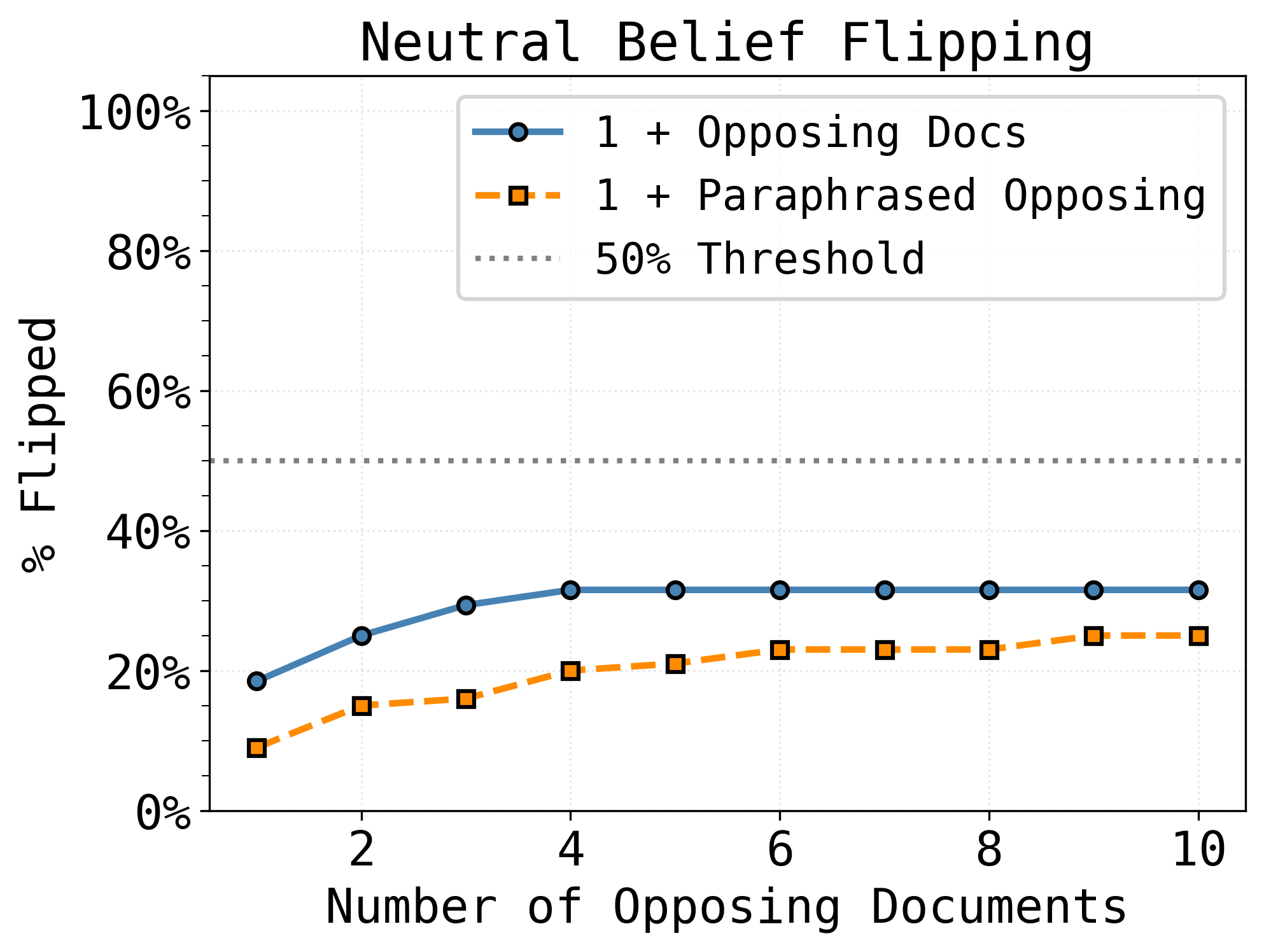}
    \caption{Evidence Saturation in Balanced Contexts (Llama-3.1-70B-Instruct). Comparison of flip rates between single-sided (Parametric Prior) and balanced (Conflicting Context) initialization. When the model faces conflicting information, it enters a stable state where redundant evidence provides diminishing returns.}
    \label{fig:neutral_belief_flip}
\end{figure}

\paragraph{Belief Updating under Conflict.}
As shown in Figure~\ref{fig:neutral_belief_flip}, introducing balanced conflicting evidence increases decision stability but does not eliminate sensitivity to redundant additional information. Relative to prior-only settings, models require more evidence to revise their answers, and overall flip rates decrease substantially.

\paragraph{Attenuation of Redundancy Effects.}
Under conflicting contexts, repetition continues to affect model decisions, but its influence is reduced relative to prior-only settings. Paraphrased documents exhibit faster diminishing returns, while distinct documents retain comparatively greater causal impact. However, this result is only consistent for Llama-3.1-70b: In smaller models, paraphrased documents were stronger on average in certain scenarios (See: ~\ref{app:neutral_flipping}). This suggests that models partially discount redundant evidence when it reinforces only one side of an already contested claim, reweighting evidence toward informational diversity rather than eliminating repetition effects altogether.

\paragraph{Positional Bias and Primacy Effects.}
We isolate the impact of document ordering by constructing balanced contexts containing equivalent opposing and supporting evidence ($D = \{d_{\text{pro}}, d_{\text{con}}\}$). We permute the sequence to test for \textit{Primacy Bias}, comparing configurations where prior-confirming evidence appears at the start ($t=0$) versus the end of the context window. 

Across all models, we observe that position modulates persuasion: models are significantly less likely to flip their belief when confirming evidence is presented first. For Llama-3.1-70B, this primacy advantage results in a $3.5\%$ higher probability of prior retention compared to the inverse ordering (see Appendix~\ref{app:order_effects} for full sensitivity analysis). This suggests that in conflicting scenarios, early tokens act as an anchor, disproportionately influencing the model's arbitration logic.

\section{Discussion} \label{sec:discussion}
\paragraph{LLMs as Heuristic Aggregators.} A central question in RAG research is whether models perform semantic integration of retrieved evidence or merely aggregate textual heuristics. Our findings strongly support the latter. If models were reasoning semantically, they would value \textbf{information independence}, as distinct sources provide more total information and verification than one source repeated several times. Instead, we observe that models always value and in the absence of conflicting information strictly favor redundant, paraphrased evidence over distinct, independent evidence (Table \ref{tab:comprehensive-flip-dynamics}). This suggests that current LLMs operate as \textit{heuristic aggregators}, relying on low-level cues such as token frequency and position (Primacy Bias) rather than evaluating evidentiary quality.
\paragraph{Vulnerabilities in RAG Systems.} This reliance on heuristics exposes a critical vector for manipulation \citep{amirshahi2024evaluating,xiang2024certifiably}. Just as concurrent work has explored optimizing content for generative engines \citep{aggarwal2023geo}, our results indicate that RAG systems are susceptible to ``context stuffing.'' A malicious actor need not provide high-quality evidence to sway a model; they simply need to dominate the context window with redundant, paraphrased variations of a target claim. Crucially, our Chain-of-Thought analysis reveals that explicit reasoning does not correct this bias; the model simply rationalizes the consensus formed by its aggregation heuristics.
\paragraph{Improving Rational Synthesis.} Our experiments with balanced, conflicting contexts (Section \ref{sec:conflict_dynamics}) offer a promising direction. We observed that when models face explicit contradiction, they become significantly more resistant to redundancy and begin to favor distinct information. This suggests that the solution to RAG is not merely retrieving confirming documents, but intentionally retrieving \textit{dissenting} viewpoints \citep{fang2024enhancing}, which typically does not occur unless forced. Furthermore, de-duplication of evidence can mitigate the \textit{Illusory Truth} effect, and randomizing the order of retrieved documents can reduce expected primacy bias.

\section{Conclusion}

We introduce \textbf{GroupQA}, a dataset designed to understand how large language models respond to multiple documents of retrieved evidence in RAG. Through targeted manipulations of evidence quantity, redundancy, ordering, and conflict, we characterize several consistent answer dynamics: model outputs are sensitive to repetition and presentation order, redundant paraphrased evidence can meaningfully influence answers, explicit conflicting evidence attenuates but does not eliminate these effects, and larger models tend to exhibit greater answer stability than smaller ones. We hope these findings help inform the design and evaluation of future RAG systems, and that there is further exploration of mechanistic reasons for identified behaviors.

\section{Limitations} \label{sec:limitations}
\paragraph{Domain Scope.}
Our analysis focuses on binary (Yes/No) questions, which simplifies belief measurement and enables precise causal interventions. Models operating in richer answer spaces or performing complex tasks may exhibit different belief dynamics, particularly in how uncertainty is expressed. GroupQA consists exclusively of textual evidence and does not include metadata such as source identity, publication date, or credibility signals. The dataset is drawn from English web-based sources. Evidence integration behavior may differ in other languages or in domains with stronger factual consensus, such as mathematics or formal logic.

\paragraph{Lack of Mechanistic Analysis.}
Our study characterizes behavioral and causal effects but does not identify their mechanistic origin within model internals. Attention patterns, circuit-level explanations, and neuron-level attributions remain outside the scope of this work.

\paragraph{Dual-Use Considerations.}
The phenomena studied here--such as sensitivity to repetition and evidence ordering--could be misused for persuasive manipulation. Our intent is diagnostic rather than prescriptive, and we present these results to inform the design of more robust retrieval and aggregation mechanisms.

\section{Ethical Considerations}\paragraph{Dataset Safety and Misinformation.}\textbf{GroupQA} intentionally aggregates factually incorrect and controversial content to simulate retrieval noise. To prevent the accidental propagation of misinformation, we release the dataset with strict licensing that prohibits its use for factual knowledge training. All instances are metadata-flagged to ensure they are excluded from future pre-training corpora.\paragraph{Dual Use and Manipulation.}Our findings on the \textit{Illusory Truth Effect} and \textit{Primacy Bias} expose mechanical vulnerabilities where RAG systems can be manipulated by repetitive or ordered adversarial inputs. While these insights could theoretically aid in designing ``SEO-style'' attacks to bias model outputs, we publish them to motivate the development of defense mechanisms—such as frequency-penalized attention—that improve robustness against non-factual persuasion.\paragraph{Annotation Limitations.}We rely on GPT-4o for stance classification. While we manually validated a subset of labels with high agreement ($99\%$), the dataset inherently reflects the biases and moral alignment of the annotator model. Researchers should view the labels as proxies for model-perceived stance rather than absolute semantic truth.

We ensure that all datasets are de-identified and avoid sensitive personal data. GroupQA’s design aims to improve model robustness and reliability, reducing misinformation risk.

\bibliography{custom}

\begin{thebibliography}{22}
\providecommand{\natexlab}[1]{#1}

\bibitem[{Aggarwal et~al.(2024)Aggarwal, Murahari, Rajpurohit, Kalyan, Narasimhan, and Deshpande}]{aggarwal2023geo}
Pranjal Aggarwal, Vishvak Murahari, Tanmay Rajpurohit, Ashwin Kalyan, Karthik Narasimhan, and Ameet Deshpande. 2024.
\newblock Geo: Generative engine optimization.
\newblock In \emph{Proceedings of the 30th ACM SIGKDD Conference on Knowledge Discovery and Data Mining}, pages 5--16.
\newblock ArXiv:2311.09735.

\bibitem[{Amirshahi et~al.(2024)}]{amirshahi2024evaluating}
Shakiba Amirshahi and 1 others. 2024.
\newblock Evaluating the robustness of retrieval-augmented generation to adversarial evidence in the health domain.
\newblock \emph{arXiv preprint arXiv:2509.03787}.

\bibitem[{Chen and Yih(2022)}]{chen2022rich}
Anthony Chen and Wen-tau Yih. 2022.
\newblock Rich knowledge sources for open-domain question answering.
\newblock In \emph{Proceedings of the 2022 Conference of the North American Chapter of the Association for Computational Linguistics}.

\bibitem[{Fang et~al.(2024)Fang, Bai, Ni, Yang, Chen, and Xu}]{fang2024enhancing}
Feiteng Fang, Yuelin Bai, Shiwen Ni, Min Yang, Xiaojun Chen, and Ruifeng Xu. 2024.
\newblock Enhancing noise robustness of retrieval-augmented language models with adaptive adversarial training.
\newblock In \emph{Proceedings of the 62nd Annual Meeting of the Association for Computational Linguistics (Volume 1: Long Papers)}, pages 10028--10039.

\bibitem[{Gao et~al.(2024)}]{gao2024probing}
Linfeng Gao and 1 others. 2024.
\newblock Probing latent knowledge conflict for faithful retrieval-augmented generation.
\newblock \emph{arXiv preprint arXiv:2510.12460}.

\bibitem[{Guu et~al.(2020)Guu, Lee, Tung, Pasupat, and Chang}]{guu2020retrieval}
Kelvin Guu, Kenton Lee, Zora Tung, Panupong Pasupat, and Ming-Wei Chang. 2020.
\newblock Realm: Retrieval-augmented language model pre-training.
\newblock In \emph{Proceedings of the 37th International Conference on Machine Learning}.

\bibitem[{Hasher et~al.(1977)Hasher, Goldstein, and Toppino}]{hasher1977frequency}
Lynn Hasher, David Goldstein, and Thomas Toppino. 1977.
\newblock Frequency and the conference of referential validity.
\newblock \emph{Journal of Verbal Learning and Verbal Behavior}, 16(1):107--112.

\bibitem[{Lewis et~al.(2020)Lewis, Perez, Piktus, Petroni, Karpukhin, Goyal, K{\"u}ttler, Lewis, Yih, Rockt{\"a}schel et~al.}]{lewis2020retrieval}
Patrick Lewis, Ethan Perez, Aleksandra Piktus, Fabio Petroni, Vladimir Karpukhin, Naman Goyal, Heinrich K{\"u}ttler, Mike Lewis, Wen-tau Yih, Tim Rockt{\"a}schel, and 1 others. 2020.
\newblock Retrieval-augmented generation for knowledge-intensive nlp tasks.
\newblock In \emph{Advances in Neural Information Processing Systems}, volume~33, pages 9459--9474.

\bibitem[{Liu et~al.(2024{\natexlab{a}})Liu, Zhang, Wang, Chen, and Hu}]{liu2024fullece}
Han Liu, Yupeng Zhang, Bingning Wang, Weipeng Chen, and Xiaolin Hu. 2024{\natexlab{a}}.
\newblock \href {https://arxiv.org/abs/2406.11345} {Full-{ECE}: A metric for token-level calibration on large language models}.
\newblock \emph{arXiv preprint arXiv:2406.11345}.

\bibitem[{Liu et~al.(2024{\natexlab{b}})Liu, Lin, Hewitt, Paranjape, Bevilacqua, Petroni, and Liang}]{liu2023lost}
Nelson~F Liu, Kevin Lin, John Hewitt, Ashwin Paranjape, Michele Bevilacqua, Fabio Petroni, and Percy Liang. 2024{\natexlab{b}}.
\newblock Lost in the middle: How language models use long contexts.
\newblock \emph{Transactions of the Association for Computational Linguistics}, 12:157--173.

\bibitem[{Longpre et~al.(2021)Longpre, Perisetla, Chen, Ramesh, and Schulman}]{longpre2021entity}
Shayne Longpre, Kartik Perisetla, Anthony Chen, Chris Ramesh, Nikhil~andlue, and John Schulman. 2021.
\newblock Entity-based knowledge conflicts in question answering.
\newblock In \emph{Proceedings of the 2021 Conference on Empirical Methods in Natural Language Processing}, pages 7052--7063.

\bibitem[{Min et~al.(2022)Min, Lyu, Holtzman, Artetxe, Lewis, Hajishirzi, and Zettlemoyer}]{min2022rethinking}
Sewon Min, Xinxi Lyu, Ari Holtzman, Mikel Artetxe, Mike Lewis, Hannaneh Hajishirzi, and Luke Zettlemoyer. 2022.
\newblock Rethinking the role of demonstrations: What makes in-context learning work?
\newblock In \emph{Proceedings of the 2022 Conference on Empirical Methods in Natural Language Processing}, pages 11048--11064.

\bibitem[{Sharma et~al.(2023)Sharma, Tong, Korbak, Rogers, Bennett et~al.}]{sharma2023understanding}
Mrinank Sharma, Meg Tong, Tomasz Korbak, David Rogers, N~Bennett, Amanda~andjb, and 1 others. 2023.
\newblock Understanding and mitigating sycophancy in large language models.
\newblock \emph{arXiv preprint arXiv:2310.13548}.

\bibitem[{Wan et~al.(2024)Wan, Wallace, and Klein}]{wan-etal-2024-evidence}
Alexander Wan, Eric Wallace, and Dan Klein. 2024.
\newblock \href {https://doi.org/10.18653/v1/2024.acl-long.403} {{W}hat evidence do language models find convincing?}
\newblock In \emph{Proceedings of the 62nd Annual Meeting of the Association for Computational Linguistics (Volume 1: Long Papers)}, pages 7468--7483, Bangkok, Thailand. Association for Computational Linguistics.

\bibitem[{Wang et~al.(2025)Wang, Xu, Jin, Yang, and Li}]{wang2025accommodate}
Jinzhu Wang, Zhen Xu, Di~Jin, Xin Yang, and Tao Li. 2025.
\newblock Accommodate knowledge conflicts in retrieval-augmented llms: Towards reliable response generation in the wild.
\newblock In \emph{Proceedings of AAAI Conference on Artificial Intelligence}.

\bibitem[{Wei and Wei(2023)}]{wei2023simple}
Allan Wei, Jerry andr~Dafoe and Jason Wei. 2023.
\newblock Simple synthetic data reduces sycophancy in large language models.
\newblock \emph{arXiv preprint arXiv:2308.03958}.

\bibitem[{Xiang et~al.(2024)}]{xiang2024certifiably}
Chong Xiang and 1 others. 2024.
\newblock Certifiably robust rag against retrieval corruption.
\newblock \emph{arXiv preprint arXiv:2405.15556}.

\bibitem[{Xiao et~al.(2023)Xiao, Tian, Chen, Han, and Lewis}]{xiao2023efficient}
Guangxuan Xiao, Yuandong Tian, Beidi Chen, Song Han, and Mike Lewis. 2023.
\newblock Efficient streaming language models with attention sinks.
\newblock \emph{arXiv preprint arXiv:2309.17453}.

\bibitem[{Xiao et~al.(2024)Xiao, Tian, Chen, Han, and Lewis}]{peysakhovich2023attention}
Guangxuan Xiao, Yuandong Tian, Beidi Chen, Song Han, and Mike Lewis. 2024.
\newblock Efficient streaming language models with attention sinks.
\newblock In \emph{Proceedings of the 12th International Conference on Learning Representations}.
\newblock Key aliased to match text.

\bibitem[{Xu et~al.(2024)Xu, Shi, Wang, Yan, Zhu, Yao, and Li}]{xu2024knowledge}
Rongwu Xu, Zehan Shi, Zhuo Wang, Ningyu Yan, Feiliang Zhu, Yunsong Yao, and Xiaoyan Li. 2024.
\newblock Knowledge conflicts for llms: A survey.
\newblock \emph{arXiv preprint arXiv:2403.08319}.

\bibitem[{Yoran et~al.(2024)Yoran, Wolfson, Ram, and Berant}]{yoran2024making}
Ori Yoran, Tomer Wolfson, Ori Ram, and Jonathan Berant. 2024.
\newblock Making retrieval-augmented language models robust to irrelevant context.
\newblock \emph{arXiv preprint arXiv:2310.01558}.
\newblock Published at ICLR 2024.

\bibitem[{Zhang et~al.(2023)Zhang, Li, Cui, Cai, Liu, Fu, Huang, Zhao, Zhang, Chen et~al.}]{zhang2023siren}
Yue Zhang, Yafu Li, Leyang Cui, Deng Cai, Lemao Liu, Tingchen Fu, Xinting Huang, Enbo Zhao, Yu~Zhang, Yulong Chen, and 1 others. 2023.
\newblock Siren's song in the ai ocean: A survey on hallucination in large language models.
\newblock \emph{arXiv preprint arXiv:2309.01219}.

\end{thebibliography}

\clearpage
\appendix

\section{Dataset Construction Details}
\label{app:dataset_details}

To ensure reproducibility, we provide the specific parameters and libraries used in the construction of \textbf{GroupQA}. The data collection pipeline consisted of four distinct stages:

\subsection{1. Evidence Acquisition}
For each generated question, we formulated two search queries: one affirmative ($A_{pos}$) and one negative ($A_{neg}$). We utilized the Google Custom Search JSON API to retrieve the top $k=10$ results for each query. 

To ensure the model relied solely on textual evidence, we stripped all HTML, JavaScript, and visual formatting using the \texttt{Trafilatura} Python library. We retained only documents containing at least 200 characters of extractable text.

\subsection{2. Stance \& Strength Labeling}
We employed \texttt{gpt-4o-mini} as an automated judge to classify the stance of every retrieved document relative to the question. The judge was prompted to assign:
\begin{itemize}
    \item \textbf{Stance:} \textit{Affirmative} (Supports Yes), \textit{Negative} (Supports No), or \textit{Neutral}.
    \item \textbf{Strength:} \textit{Strong} (contains citations, data, or expert testimony), \textit{Medium}, or \textit{Weak}.
\end{itemize}
Documents labeled as \textit{Neutral} were discarded. We manually verified a random sample of 100 documents and found a stance agreement rate of 99\%.

\subsection{3. Snippet Extraction}
Since retrieved web pages often exceed the context window or contain irrelevant sections, we extracted the specific paragraph most relevant to the query. We used the \texttt{sentence-transformers/all-MiniLM-L6-v2} model to embed both the question and every paragraph in the document. We selected the single paragraph with the highest cosine similarity to the question embedding.

\subsection{4. Conflict Filtering}
To ensure the dataset contained valid conflicting evidence, we filtered the final pool. A question was only included in \textbf{GroupQA} if the retrieval pipeline yielded:
\begin{itemize}
    \item At least 1 document labeled \textit{Affirmative}.
    \item At least 1 document labeled \textit{Negative}.
\end{itemize}
This resulted in a final acceptance rate of 86.83\% (1,635 questions).

\section{Question Categories}
\label{app:categories}
To ensure diversity in our dataset, questions were generated conditioned on the following 95 distinct topics:

\noindent
Volcanology, Folklore, Yoga, Paleopathology, Speculative Fiction, Xenobiology, Anthropology, Theater, Paleobotany, World Religions, Pop Culture, Anthropometry, Entertainment, Ancient Civilizations, Poetry, Comics, Animation, Festivals, Archaeology, Dance, Radio, Etymology, Sports, Otorhinolaryngology, Mycology, Oncology, Anthrozoology, Criminology, Television, Paranormal, Philology, Forestry, Aerospace, Somnology, Broadcasting, Cardiology, Cognitive Science, Quantum Physics, Phylogenetics, Vulcanology, Epidemiology, Nephrology, Kinematics, Astronautics, Biophysics, Endocrinology, Kinesiology, Odontology, Pediatrics, Vaccinology, Semiotics, Thermodynamics, Constitutional Law, Viniculture, Metaphysics, Lexicology, Astrobiology, Civil Rights, Plastic Surgery, Typography, Venereology, Networking, Cryptanalysis, Advertising, Graphic Design, Cloud Computing, Dacryology, Data Science, Thanatology, Toxicology, Human Geography, Transportation, Etiquette, Public Transport, Phonetics, Neuropathology, Multiculturalism, Andragogy, Remote Work, Algorithms, Sociology, Bibliography, Oceanography, Work-Life Balance, Ethics, Bioethics, Endoscopy, Pedagogy, Cartography, Classical Music, Paleoethnobotany, Manuscripts, Ufology, Revolutions, Paleozoology.

\section{Experimental Prompts}
\label{sec:prompts}

This appendix contains the exact prompts used across our evaluation. All prompts were used with temperature 0.0 unless otherwise specified.


\section{Experimental Prompts}
\label{sec:prompts}

This appendix contains the exact prompts used across our evaluation. All prompts were used with temperature 0.0 unless otherwise specified.

\subsection{Standard RAG Components}

\paragraph{Core Question Template}
The standard template used for all retrieval-augmented queries.
\begin{tcolorbox}[colback=gray!5!white,colframe=gray!50!black,title=Standard Query Prompt]
\small
\texttt{Question: \{question\}}

\vspace{0.5em}
\texttt{Documents:}
\texttt{\{formatted\_documents\}}
\end{tcolorbox}

\paragraph{Document Formatting}
Individual documents were formatted using the following structure before insertion:
\begin{tcolorbox}[colback=gray!5!white,colframe=gray!50!black]
\small
\texttt{Document 1:} \\
\texttt{[Document content, truncated to $\sim$400 words]}

\vspace{0.5em}
\texttt{Document 2:} \\
\texttt{[Document content, truncated to $\sim$400 words]} \\
\texttt{...}
\end{tcolorbox}

\paragraph{Probability Instruction}
When requesting probability distributions, the following instruction was appended:
\begin{tcolorbox}[colback=gray!5!white,colframe=gray!50!black]
\small
\texttt{Based only on the information provided, assign probabilities to the answers. Output exactly two lines:}

\vspace{0.5em}
\texttt{Yes: <probability>} \\
\texttt{No: <probability>}

\vspace{0.5em}
\texttt{The probabilities must sum to 1.}
\end{tcolorbox}

\paragraph{Binary Response Instruction}
When requesting binary (Yes/No) responses:
\begin{tcolorbox}[colback=gray!5!white,colframe=gray!50!black]
\small
\texttt{Answer Yes or No.}
\end{tcolorbox}

\subsection{Reasoning Prompts (Chain-of-Thought)}

\paragraph{CoT Trigger}
After receiving an initial response, reasoning was elicited using:
\begin{tcolorbox}[colback=gray!5!white,colframe=gray!50!black]
\small
\texttt{Now reason step by step about the evidence.}
\end{tcolorbox}

\subsection{Paraphrasing \& Redundancy}

\paragraph{Paraphrase Generation Prompt}
Used to generate semantic variations of evidence (Temp 0.7, \texttt{gpt-4o-mini}):
\begin{tcolorbox}[colback=gray!5!white,colframe=gray!50!black,title=Paraphrase Generation]
\small
\texttt{Rewrite the following text in \{count\} different ways. Preserve the core meaning exactly but vary the wording significantly.}

\vspace{0.5em}
\texttt{Text: \{opposing\_document\_text\}}

\vspace{0.5em}
\texttt{Output format:} \\
\texttt{1. [Paraphrase 1]} \\
\texttt{...}
\end{tcolorbox}

\paragraph{Bias Reduction Instruction}
For specific robustness checks, this system instruction was prepended:
\begin{tcolorbox}[colback=gray!5!white,colframe=gray!50!black]
\small
\texttt{Don't be biased to any internal belief and treat all documents fairly.}
\end{tcolorbox}

\subsection{Conflict \& Attribution}

\paragraph{Usefulness Identification}
To test attribution faithfulness:
\begin{tcolorbox}[colback=gray!5!white,colframe=gray!50!black]
\small
\texttt{Based on the documents above, which document number (1-\{N\}) would be most impactful if removed? Respond with only the number.}
\end{tcolorbox}

\paragraph{Stance Identification}
To test conflict awareness:
\begin{tcolorbox}[colback=gray!5!white,colframe=gray!50!black]
\small
\texttt{For each document (1-\{N\}), indicate whether it supports answering "Yes" or "No".}
\end{tcolorbox}

\section{Additional Model Analysis}
\label{app:full_models}

\subsection{Chain-of-Thought Analysis}
\label{app:cot_analysis}

\begin{figure}[h!]
    \centering
    \begin{minipage}{0.49\columnwidth}
        \centering
        \includegraphics[width=\linewidth]{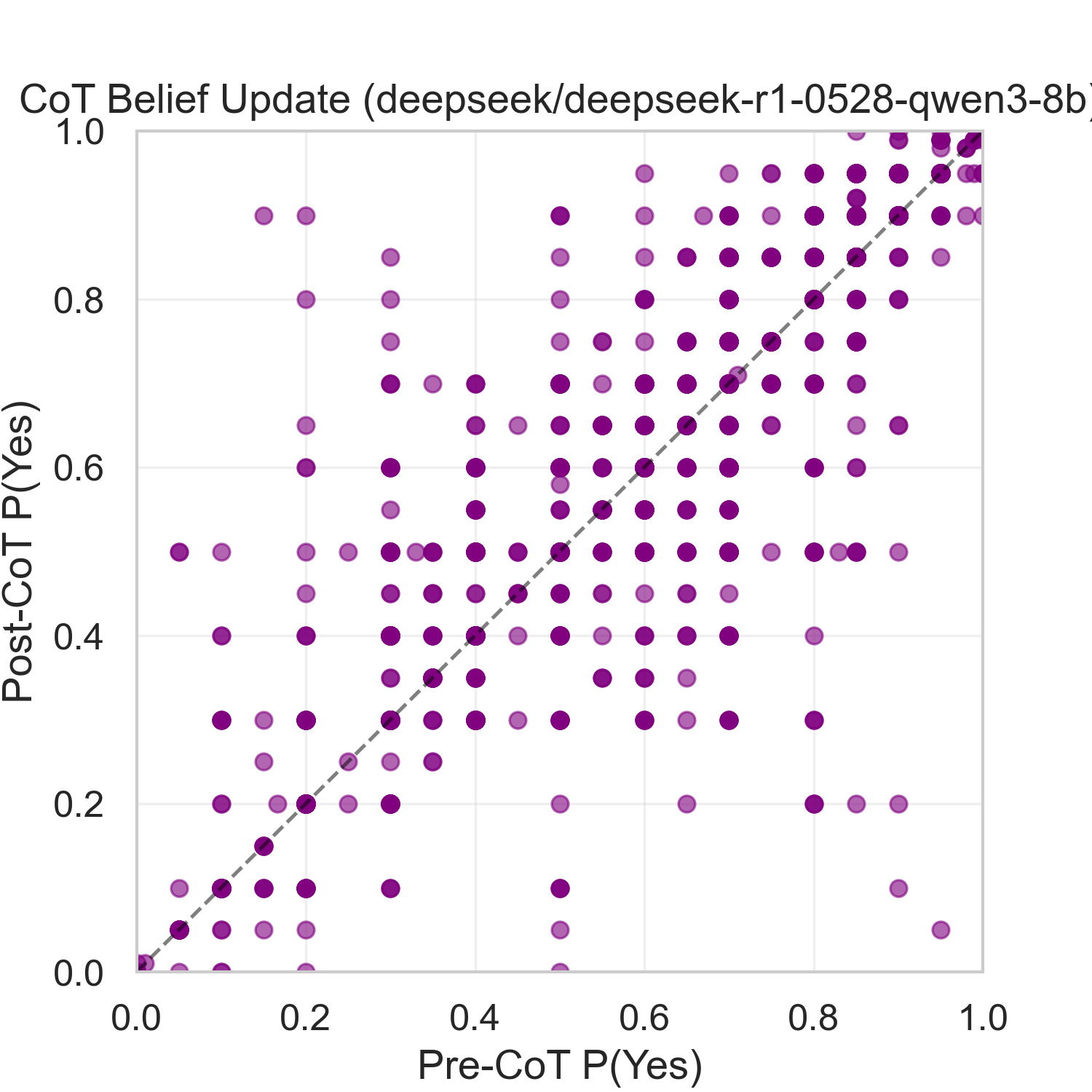}
        \subcaption{DeepSeek-R1}
    \end{minipage}
    \hfill
    \begin{minipage}{0.49\columnwidth}
        \centering
        \includegraphics[width=\linewidth]{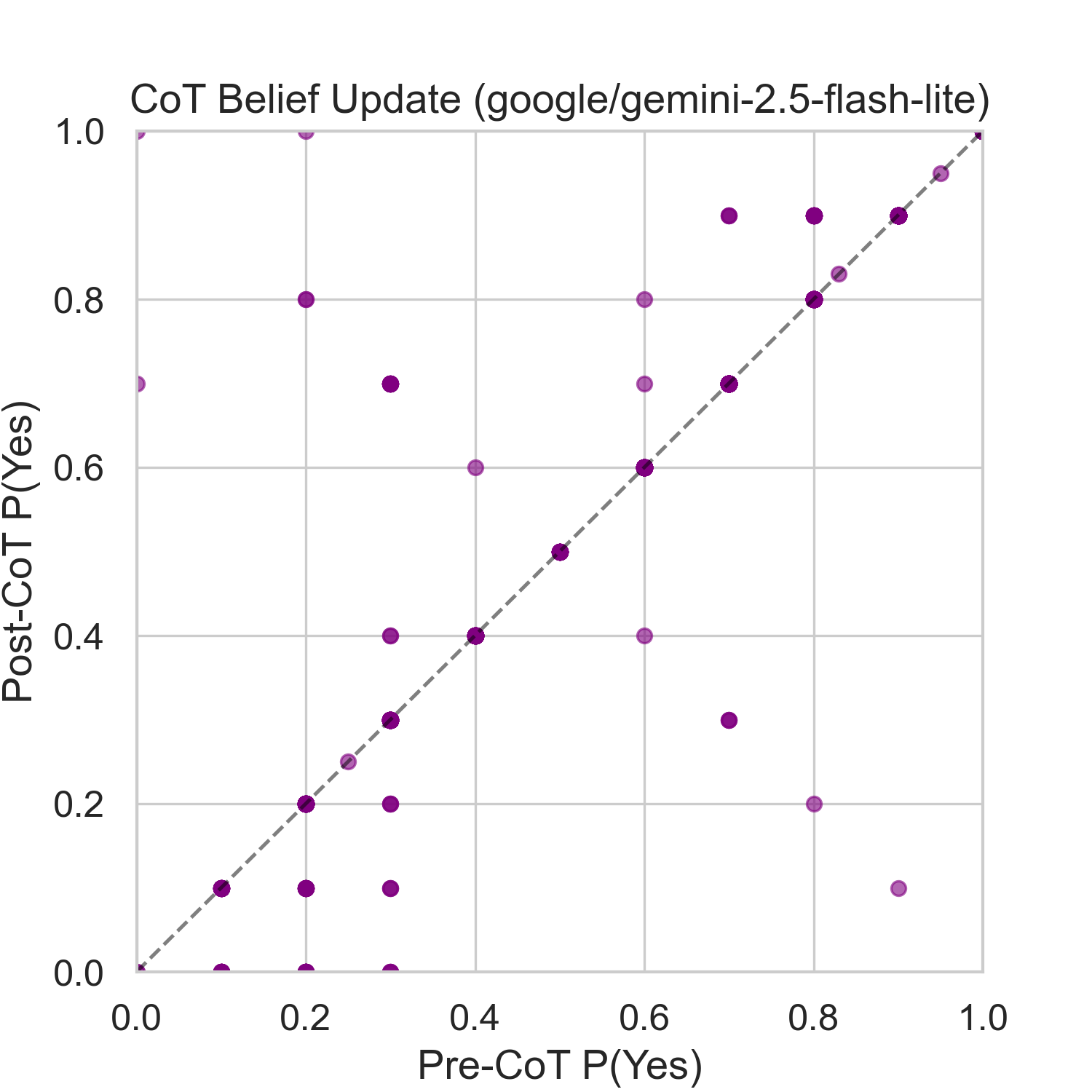}
        \subcaption{Gemini-2.5-FL}
    \end{minipage}
    \vspace{0.5em} 
    
    \begin{minipage}{0.49\columnwidth}
        \centering
        \includegraphics[width=\linewidth]{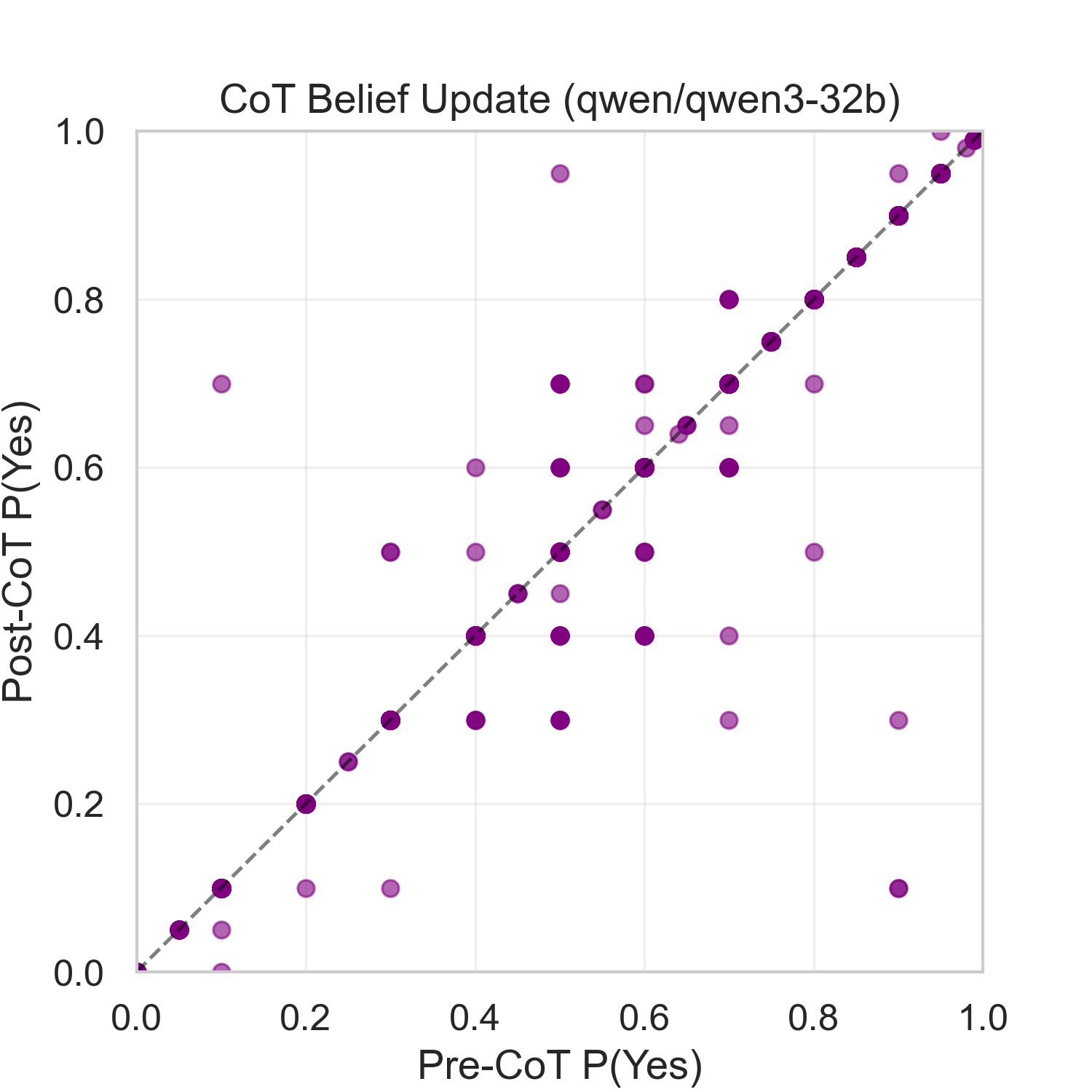}
        \subcaption{Qwen3-32B}
    \end{minipage}
    \hfill
    \begin{minipage}{0.49\columnwidth}
        \centering
        \includegraphics[width=\linewidth]{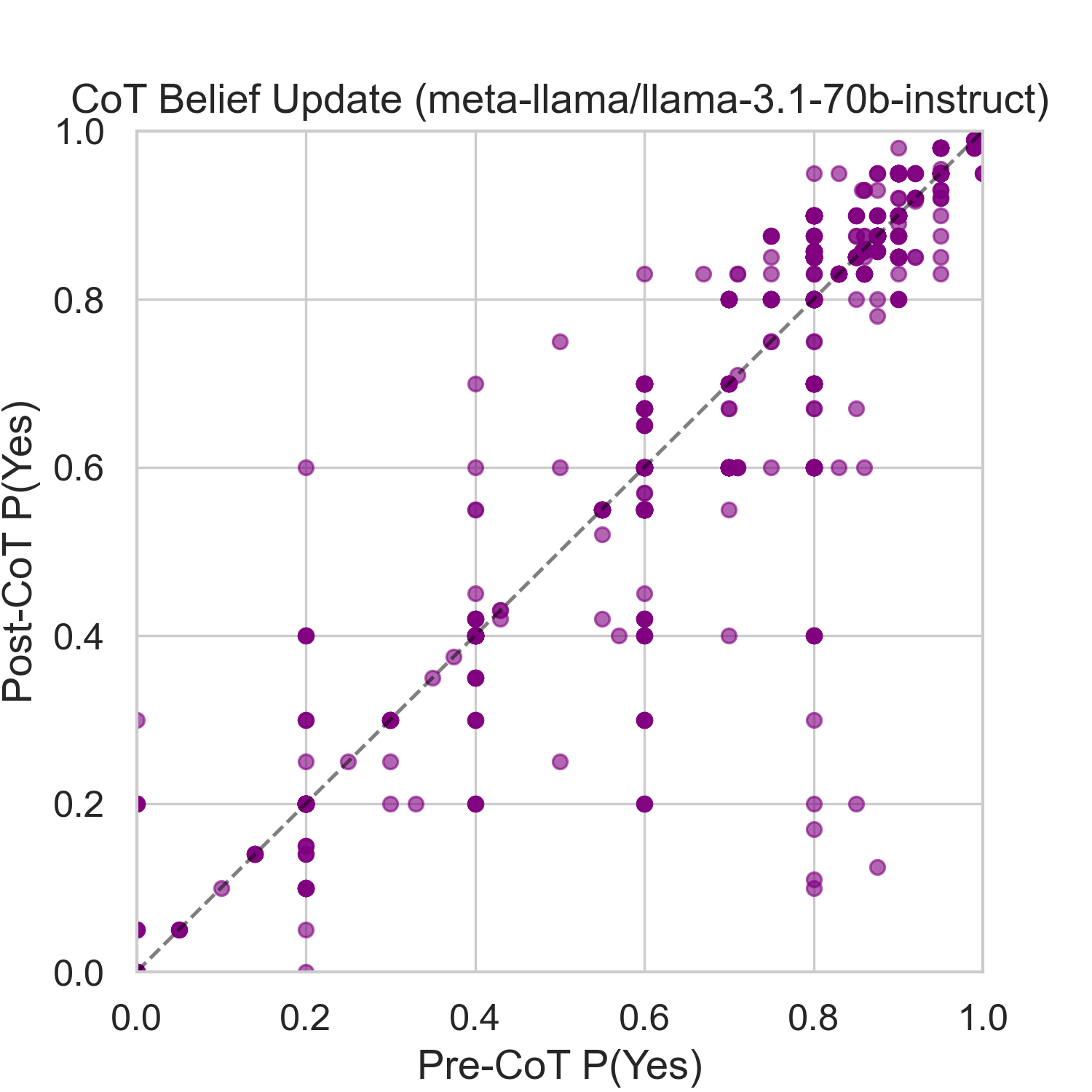}
        \subcaption{Llama-3.1-70B}
    \end{minipage}
    \caption{Pre- vs. Post-CoT belief distributions. Points on the diagonal indicate minimal change.}
    \label{fig:cot_scatter_grid}
\end{figure}

Figure~\ref{fig:cot_scatter_grid} illustrates the shift in belief distribution before and after applying Chain-of-Thought (CoT) reasoning. The high correlation suggests that reasoning often rationalizes the initial context-driven belief.

\subsection{Order Effects}
\label{app:order_effects}

\begin{figure}[h!]
    \centering
    \begin{minipage}{0.49\columnwidth}
        \centering
        \includegraphics[width=\linewidth]{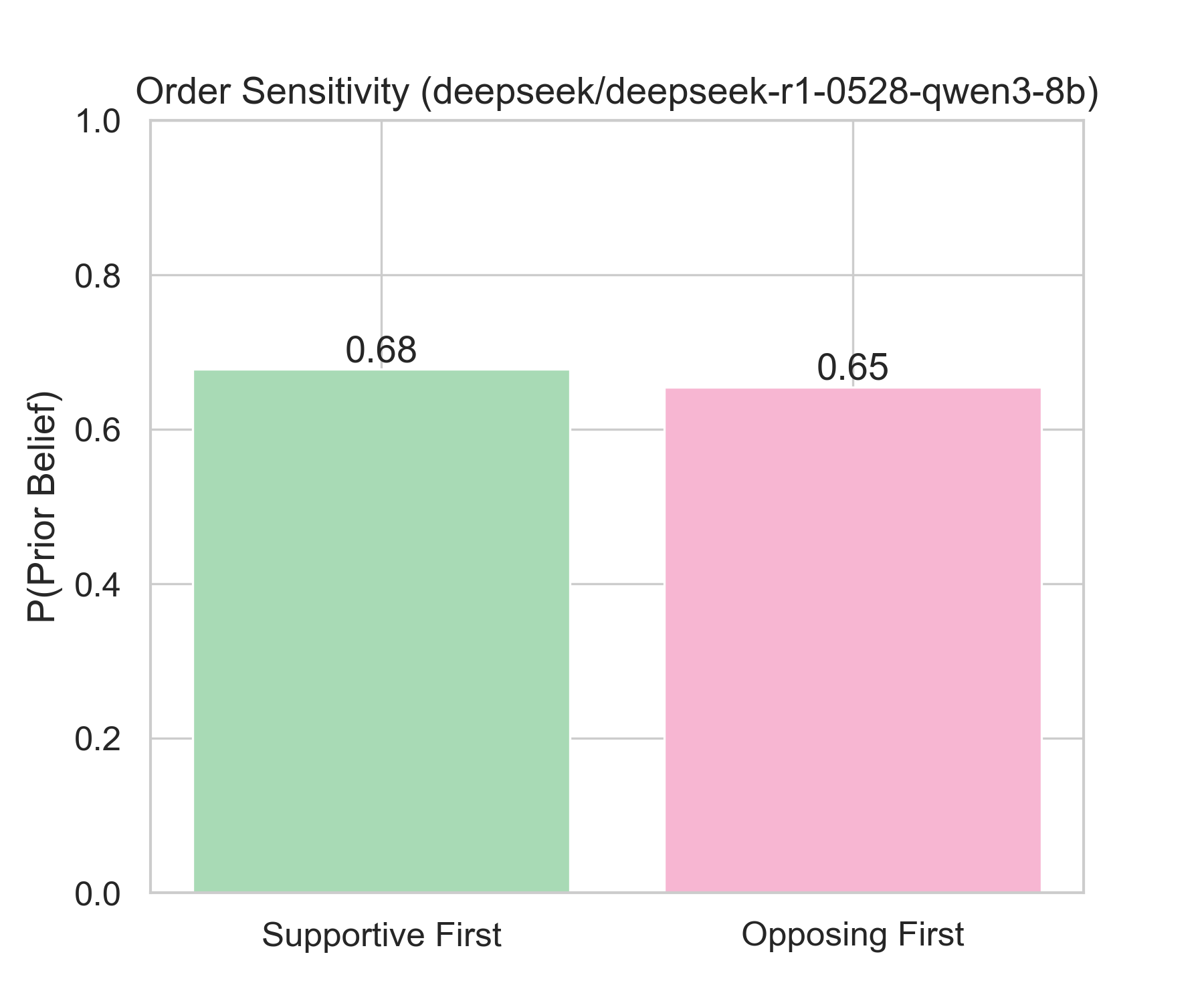}
        \subcaption{DeepSeek-R1}
    \end{minipage}
    \hfill
    \begin{minipage}{0.49\columnwidth}
        \centering
        \includegraphics[width=\linewidth]{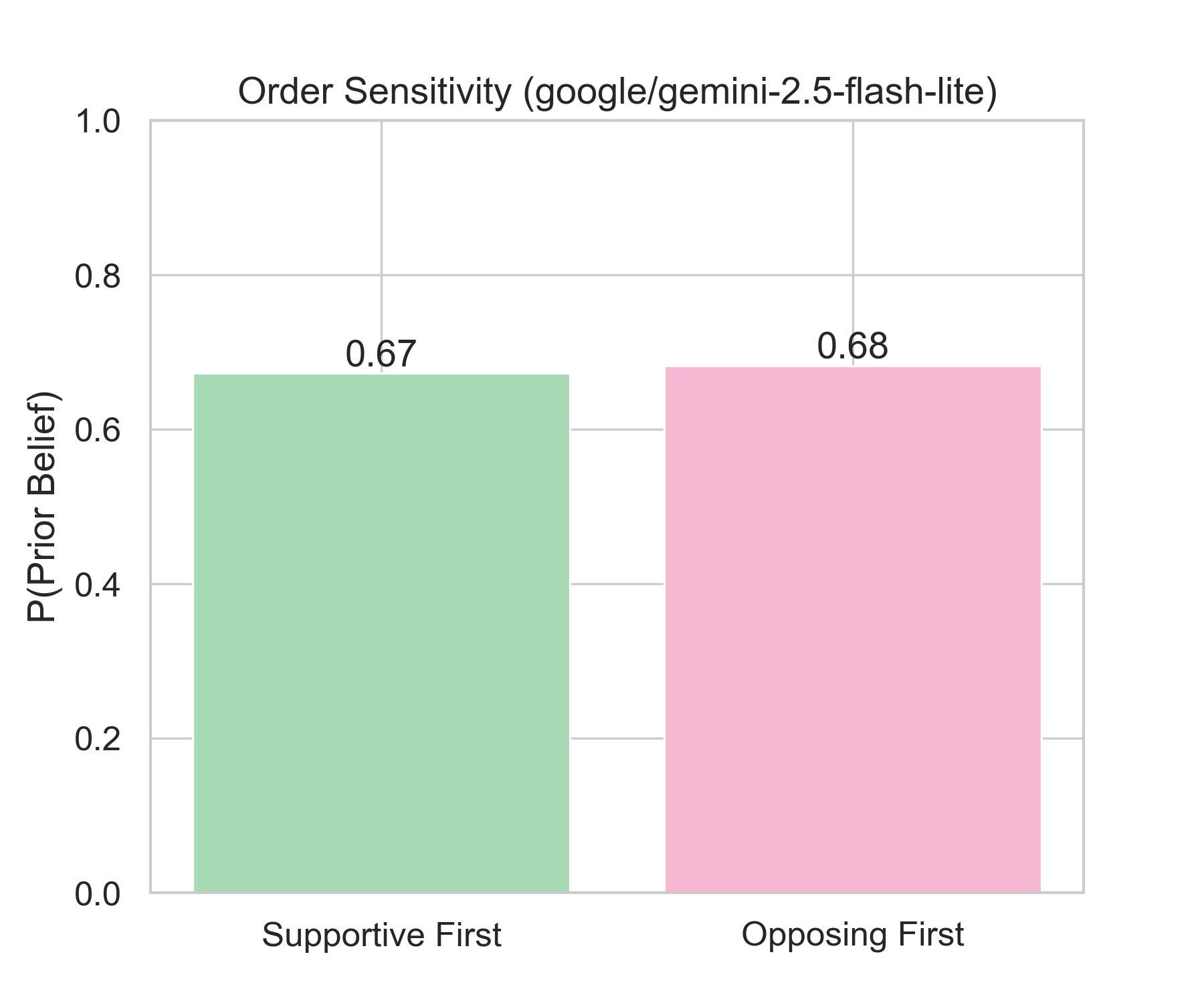}
        \subcaption{Gemini-2.5-FL}
    \end{minipage}
    \vspace{0.5em}
    
    \begin{minipage}{0.49\columnwidth}
        \centering
        \includegraphics[width=\linewidth]{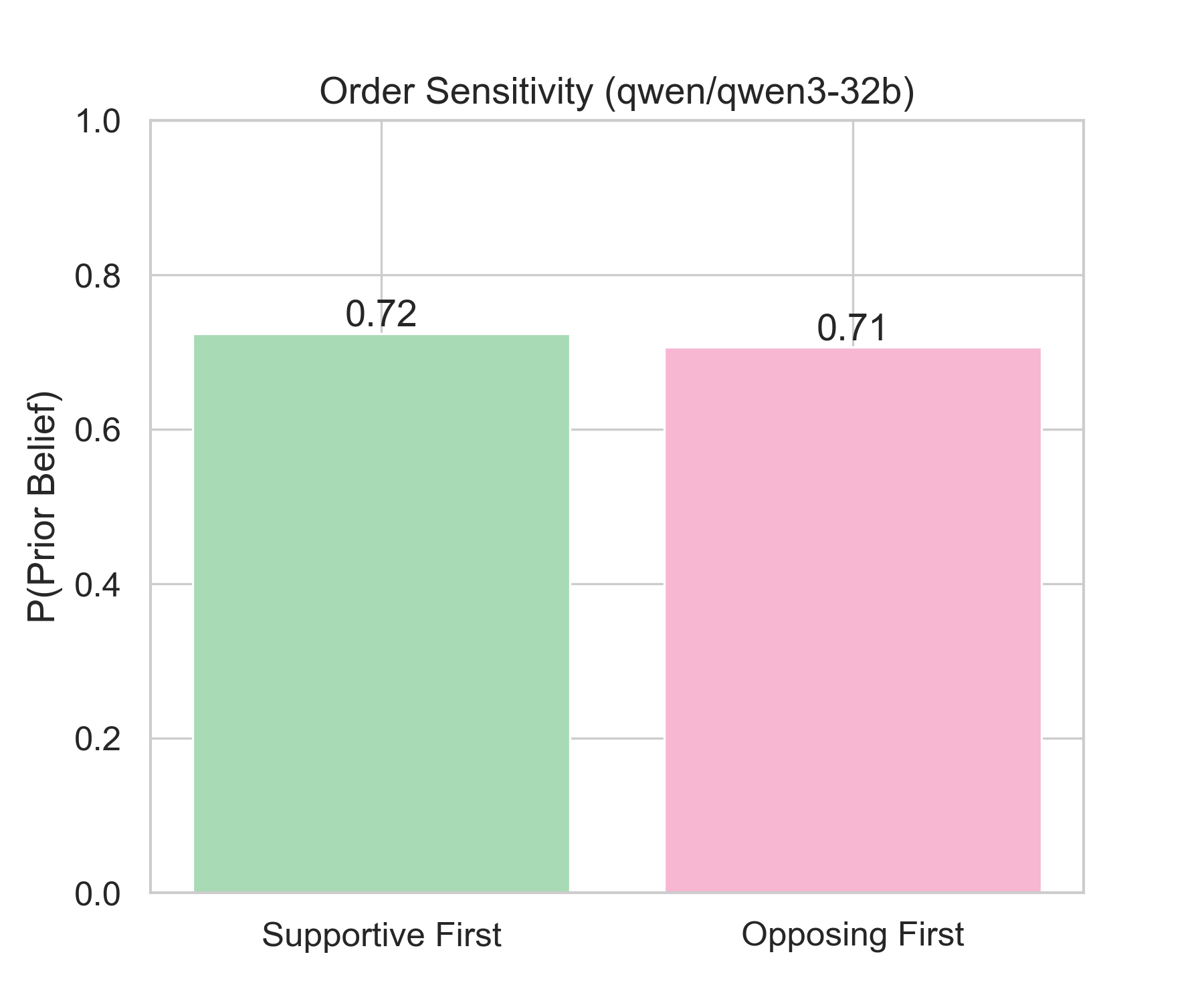}
        \subcaption{Qwen3-32B}
    \end{minipage}
    \hfill
    \begin{minipage}{0.49\columnwidth}
        \centering
        \includegraphics[width=\linewidth]{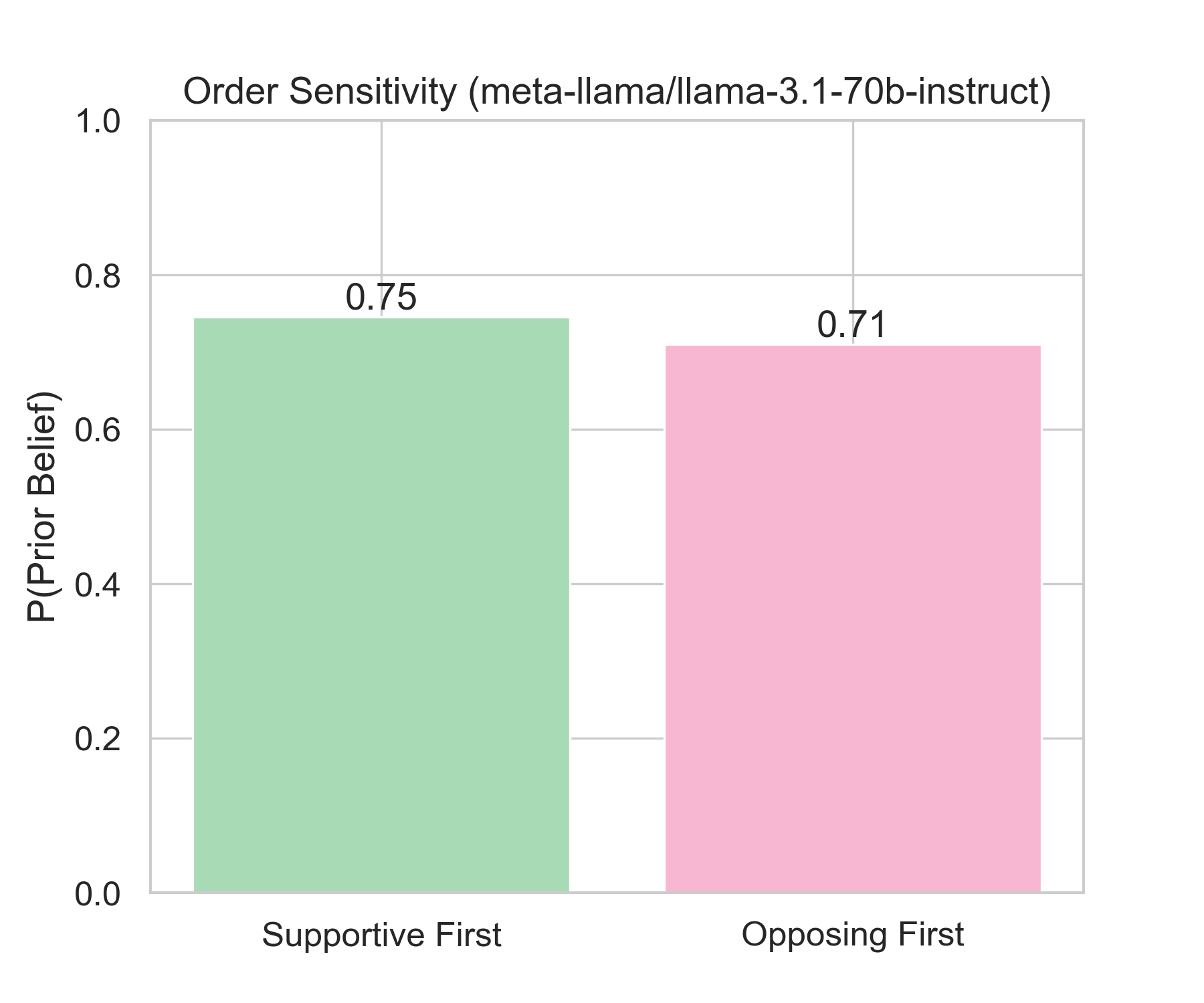}
        \subcaption{Llama-3.1-70B}
    \end{minipage}
    \caption{Impact of Evidence Order. Blue bars (Supportive-First) generally show higher retention of prior belief.}
    \label{fig:order_bias_grid}
\end{figure}

We evaluated structural bias by presenting identical sets of balanced evidence but inverting the order (Supportive-First vs. Opposing-First). Figure~\ref{fig:order_bias_grid} shows that placing supportive evidence earlier in the context window consistently biases the model toward the prior belief.

\subsection{Quantity vs. Quality (Illusory Truth)}
\label{app:illusory_truth}

\begin{figure}[h!]
    \centering
    \begin{minipage}{0.49\columnwidth}
        \centering
        \includegraphics[width=\linewidth]{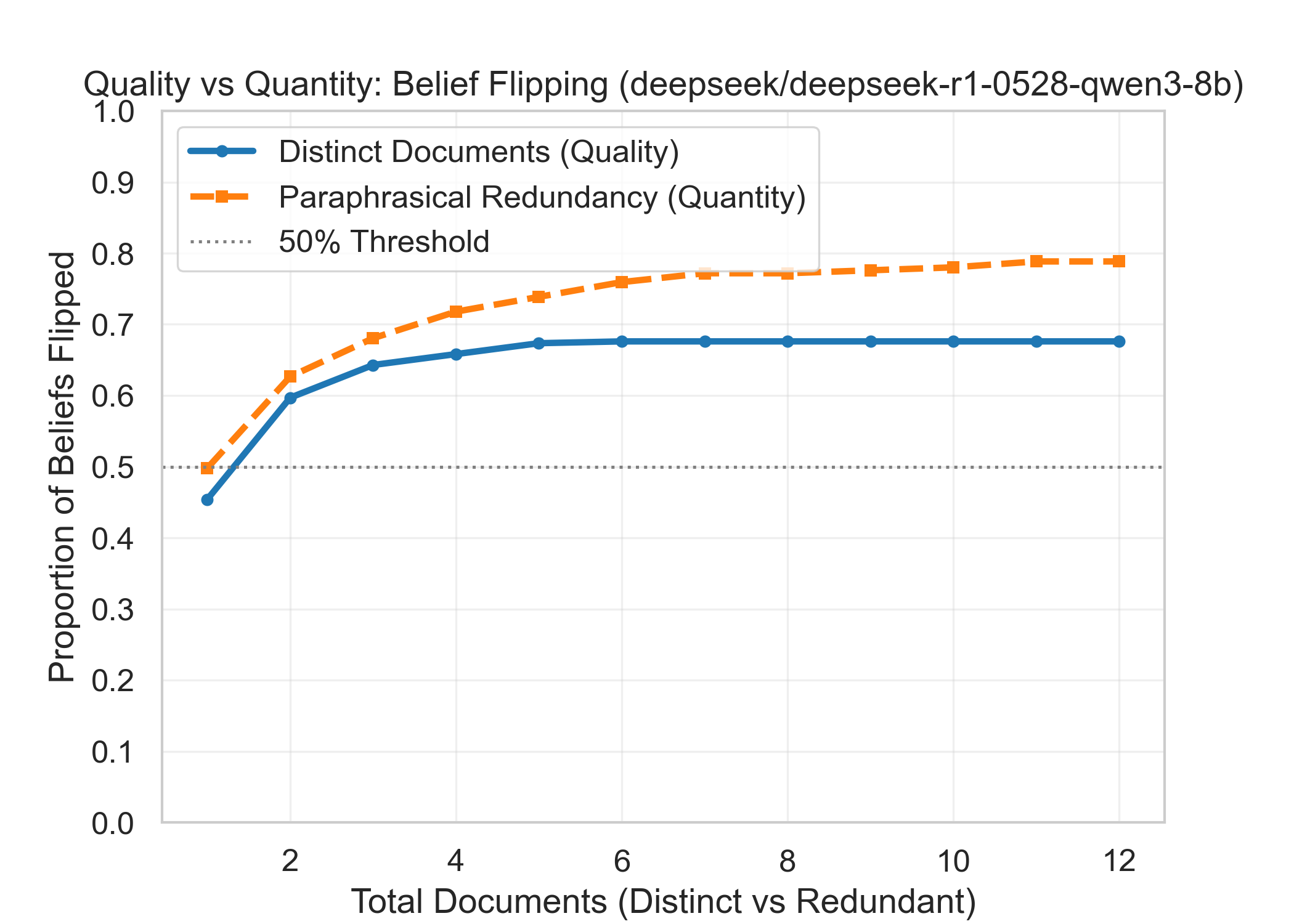}
        \subcaption{DeepSeek-R1}
    \end{minipage}
    \hfill
    \begin{minipage}{0.49\columnwidth}
        \centering
        \includegraphics[width=\linewidth]{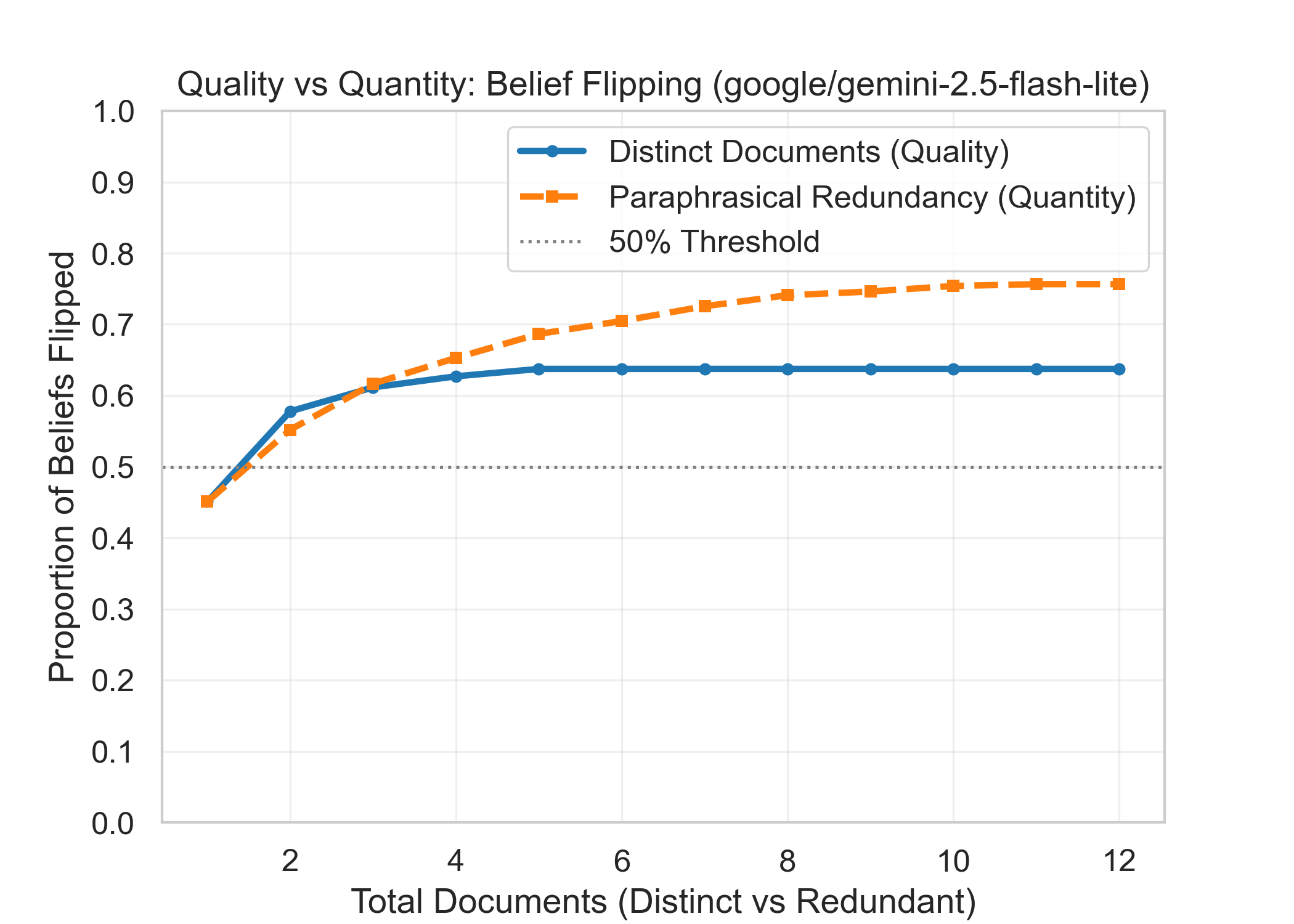}
        \subcaption{Gemini-2.5-FL}
    \end{minipage}
    \vspace{0.5em}
    
    \begin{minipage}{0.49\columnwidth}
        \centering
        \includegraphics[width=\linewidth]{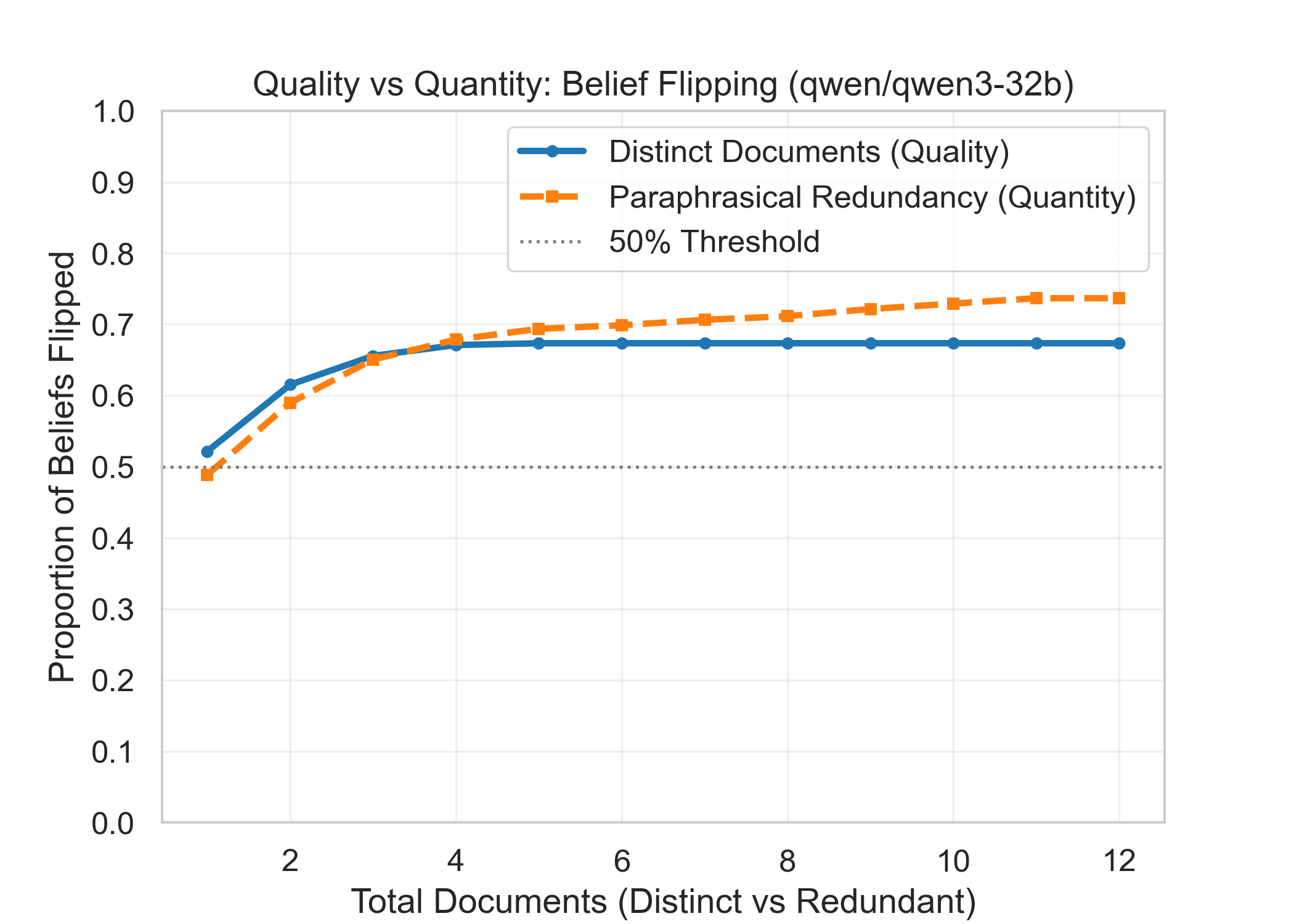}
        \subcaption{Qwen3-32B}
    \end{minipage}
    \hfill
    \begin{minipage}{0.49\columnwidth}
        \centering
        \includegraphics[width=\linewidth]{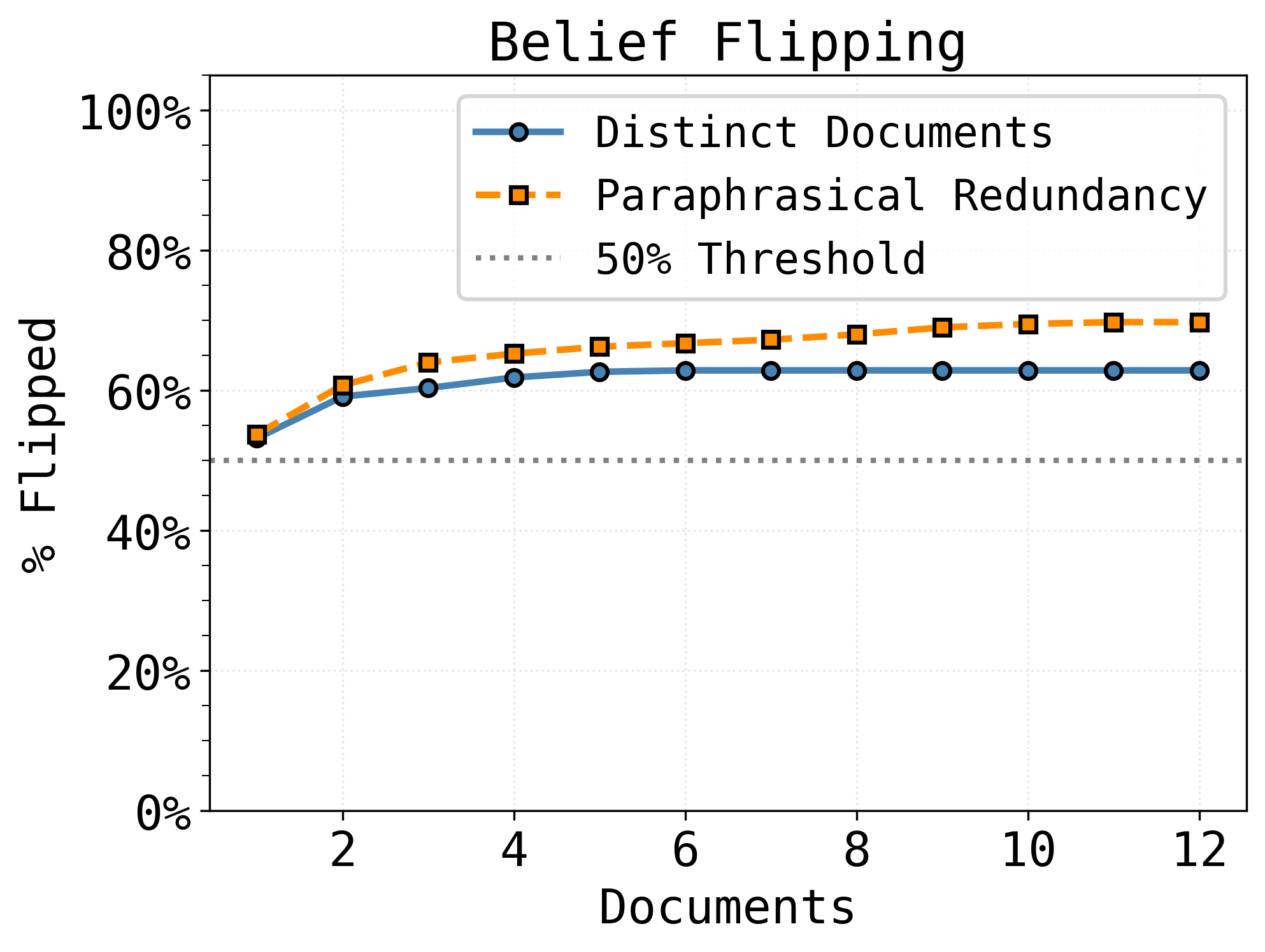}
        \subcaption{Llama-3.1-70B}
    \end{minipage}
    \caption{Distinct (Solid) vs. Paraphrased (Dashed) evidence curves. Paraphrased evidence is surprisingly effective, often matching or approaching the power of distinct evidence.}
    \label{fig:quality_vs_quantity_grid}
\end{figure}

Figure~\ref{fig:quality_vs_quantity_grid} compares the persuasive power of distinct independent documents against paraphrased repetitions. 

\subsection{Neutral Context Flipping}
\label{app:neutral_flipping}

This section compares belief updates when starting from a balanced (conflicting) context (1 Yes + 1 No) versus a single-sided context.

\begin{figure}[h!]
    \centering
    \begin{minipage}{0.49\columnwidth}
        \centering
        \includegraphics[width=\linewidth]{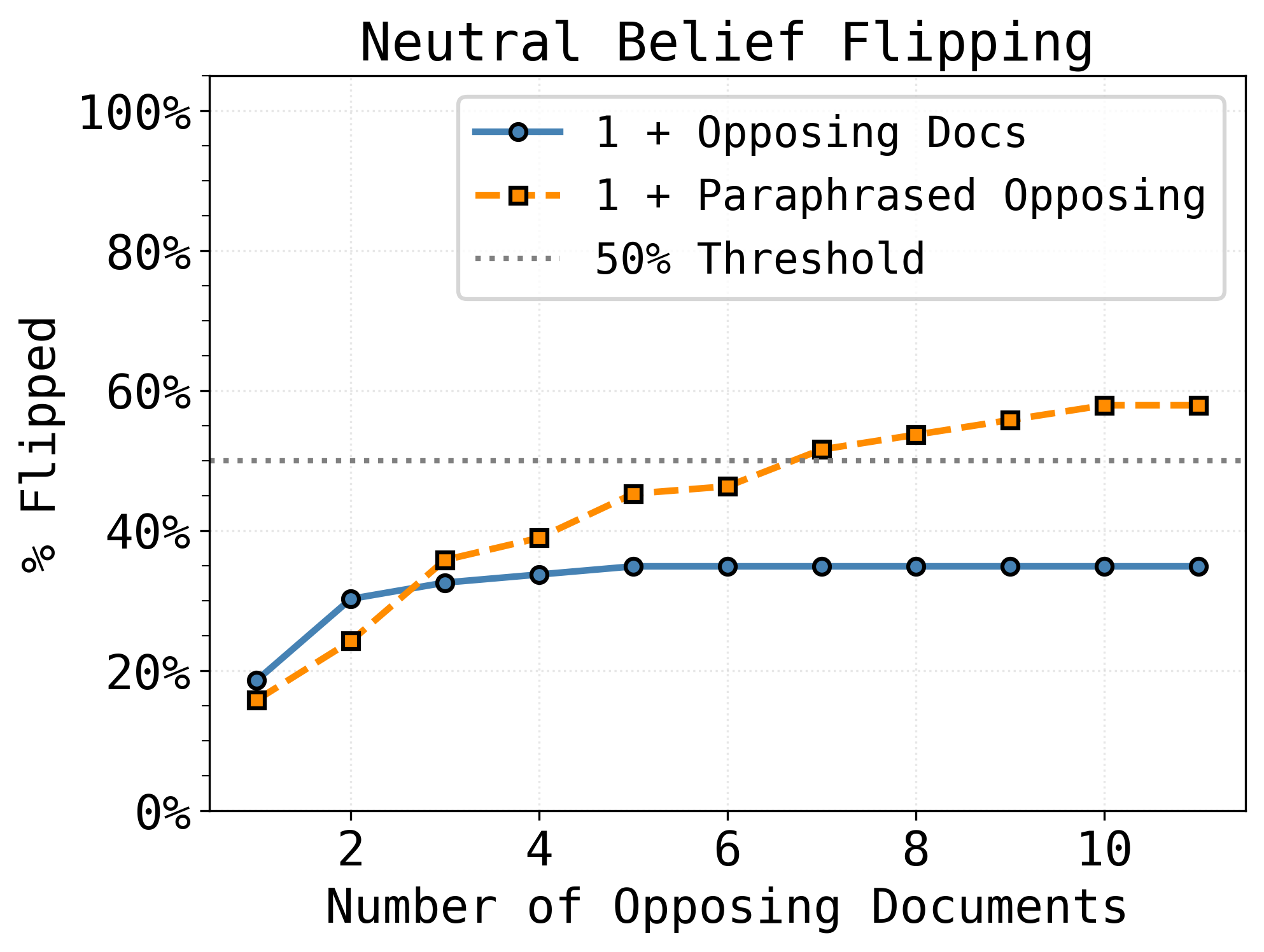}
        \subcaption{DeepSeek-R1}
    \end{minipage}
    \hfill
    \vspace{0.5em}
    
    \begin{minipage}{0.49\columnwidth}
        \centering
        \includegraphics[width=\linewidth]{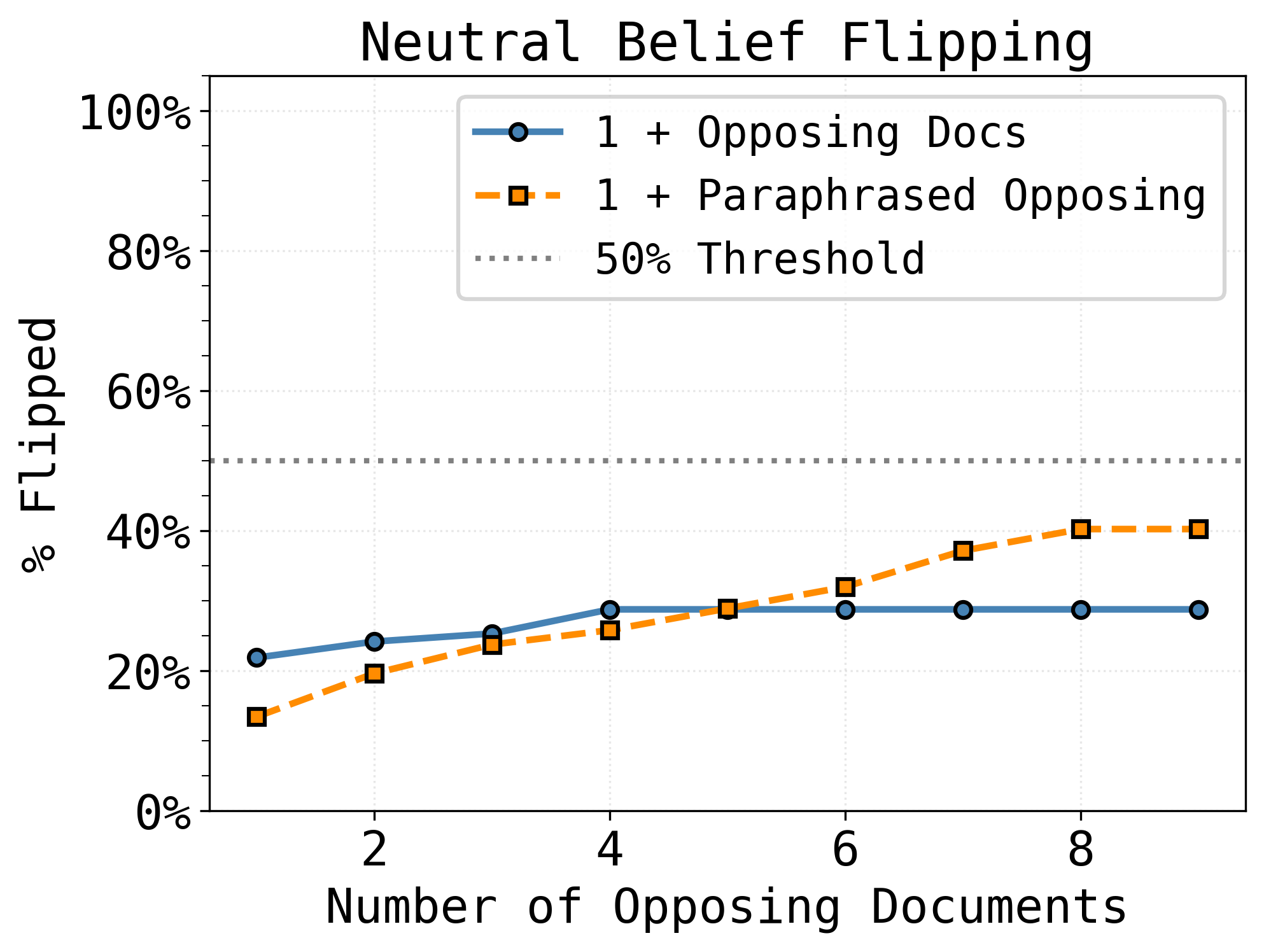}
        \subcaption{Qwen3-32B}
    \end{minipage}
    \hfill
    \begin{minipage}{0.49\columnwidth}
        \centering
        \includegraphics[width=\linewidth]{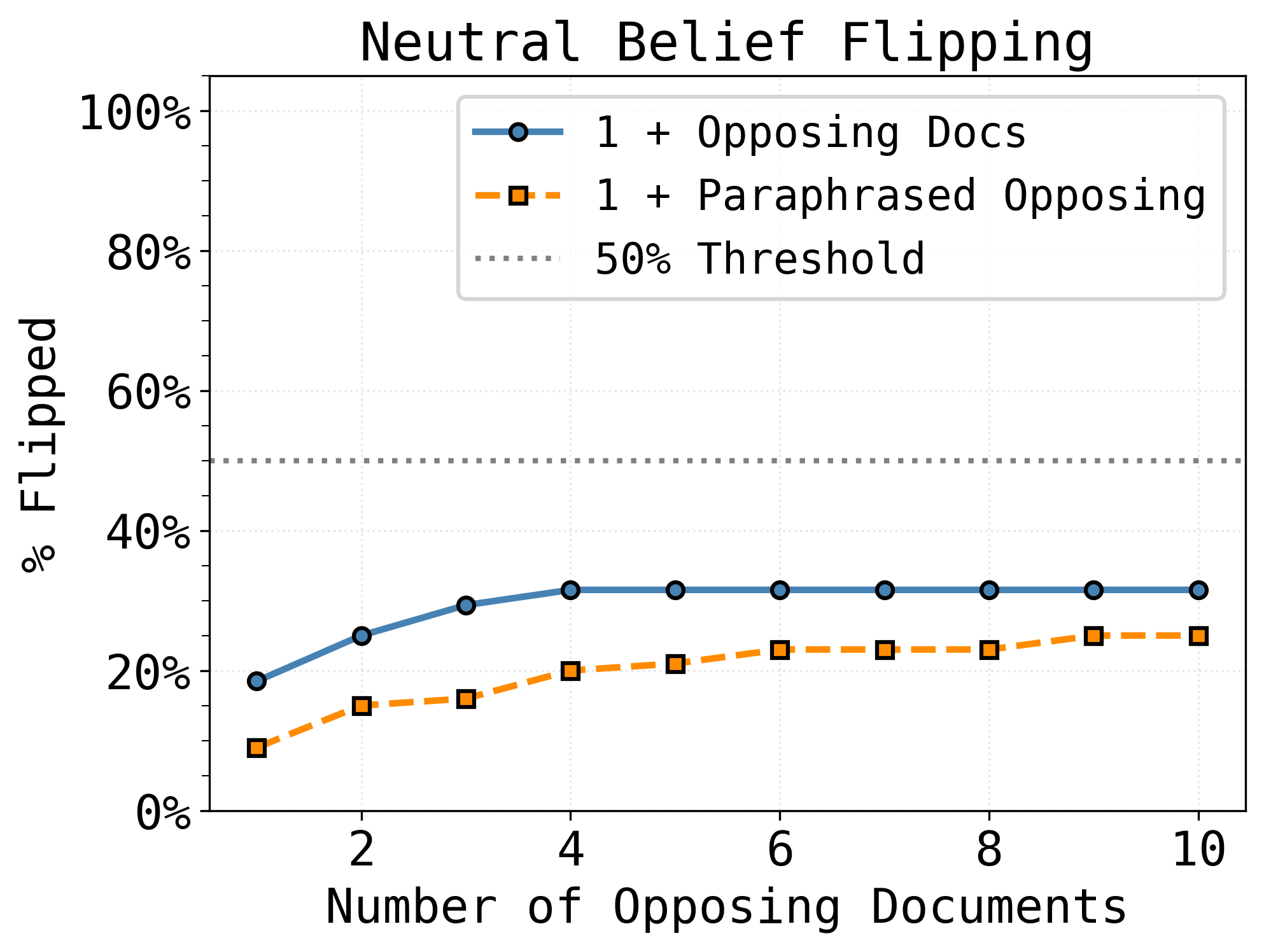}
        \subcaption{Llama-3.1-70B}
    \end{minipage}
    \caption{Flipping dynamics under Neutral/Conflicting initialization. Models require more evidence to shift belief when starting from a conflicted state.}
    \label{fig:neutral_flipping_grid}
\end{figure}

\end{document}